\documentclass[sigconf]{acmart}
\AtBeginDocument{%
  }

\usepackage{algorithm}
\usepackage{algpseudocode}

\acmYear{2026}\copyrightyear{2026}
\setcopyright{rightsretained}
\acmConference[ACM CAIS '26]{ACM Conference on AI and Agentic Systems}{May 26--29, 2026}{San Jose, CA, USA}
\acmBooktitle{ACM Conference on AI and Agentic Systems (ACM CAIS '26), May 26--29, 2026, San Jose, CA, USA}
\acmDOI{10.1145/3786335.3813155}
\acmISBN{979-8-4007-2415-2/26/05}

\newcommand{\up}{\ensuremath{\,{\scriptstyle\uparrow}}}
\newcommand{\dn}{\ensuremath{\,{\scriptstyle\downarrow}}}

\usepackage{multirow}

\begin{document}

\title[FORGE: Self-Evolving Agent Memory With No Weight Updates]{FORGE: Self-Evolving Agent Memory With No Weight Updates via Population Broadcast}

\author{Igor Bogdanov}
\email{igorbogdanov@cmail.carleton.ca}
\orcid{0009-0008-6606-189X}
\affiliation{%
  \institution{Carleton University}
  \city{Ottawa}
  \state{Ontario}
  \country{Canada}
}

\author{Chung-Horng Lung}
\email{chlung@sce.carleton.ca}
\orcid{0000-0002-5662-490X}
\affiliation{%
  \institution{Carleton University}
  \city{Ottawa}
  \state{Ontario}
  \country{Canada}
}

\author{Thomas Kunz}
\email{tkunz@sce.carleton.ca}
\orcid{0000-0002-6241-778X}
\affiliation{%
  \institution{Carleton University}
  \city{Ottawa}
  \state{Ontario}
  \country{Canada}
}

\author{Jie Gao}
\email{jie.gao6@carleton.ca}
\orcid{0000-0001-6095-2968}
\affiliation{%
  \institution{Carleton University}
  \city{Ottawa}
  \state{Ontario}
  \country{Canada}
}

\author{Adrian Taylor}
\email{Adrian.Taylor@forces.gc.ca}
\orcid{0000-0002-3785-6270}
\affiliation{%
  \institution{Defence R\&D Canada}
  \city{Ottawa}
  \state{Ontario}
  \country{Canada}
}

\author{Marzia Zaman}
\email{Marzia@cistel.com}
\orcid{0000-0002-0610-0470}
\affiliation{%
  \institution{Cistel Technology}
  \city{Ottawa}
  \state{Ontario}
  \country{Canada}
}

\renewcommand{\shortauthors}{Bogdanov et al.}

\begin{abstract}
    Can LLM agents improve decision-making through self-generated memory without gradient updates? We propose FORGE (Failure-Optimized Reflective Graduation and Evolution), a staged, population-based protocol that evolves prompt-injected natural-language memory for hierarchical ReAct agents. FORGE wraps a Reflexion-style inner loop, where a dedicated reflection agent (using the same underlying LLM, no distillation from a stronger model) converts failed trajectories into reusable knowledge artifacts: textual heuristics (\textsc{Rules}), few-shot demonstrations (\textsc{Examples}), or both (\textsc{Mixed}), with an outer loop that propagates the best-performing instance's memory to the population between stages and freezes converged instances via a graduation criterion. We evaluate on CybORG CAGE-2, a stochastic network-defense POMDP at a 30-step horizon against the B\_line attacker, where all four tested LLM families (Gemini-2.5-Flash-Lite, Grok-4-Fast, Llama-4-Maverick, Qwen3-235B) exhibit strongly negative, heavy-tailed zero-shot rewards. Compared against both a zero-shot baseline and a Reflexion baseline (isolated single-stream learning), FORGE improves average evaluation return by 1.7-7.7$\times$ over zero-shot and by 29-72\% over Reflexion in all 12 model-representation conditions, reducing major-failure rates (below $-100$) to as low as $\sim$1\%. We find that (1) population broadcast is the critical mechanism, with a no-graduation ablation confirming that broadcast carries the performance gains while graduation primarily saves compute; (2) \textsc{Examples} achieves the strongest returns for three of four models, while \textsc{Rules} offers the best cost-reliability profile with $\sim$40\% fewer tokens; and (3) weaker baseline models benefit disproportionately, suggesting FORGE may mitigate capability gaps rather than amplify strong models. All evidence is confined to CAGE-2 B\_line; cross-family findings are directional evidence.
\end{abstract}



%% (No teaser figure for this paper)

\maketitle

\section{Introduction}

Large language models (LLMs) can act as general-purpose reasoning engines for sequential decision-making with ability to self-reflect and improve when embedded in agentic scaffolds such as ReAct \citep{yao2023react}, Reflexion \citep{shinn2023reflexion}, and Voyager \citep{wang2023voyager}. Yet most such agents remain single-episode systems: they reason and act within an episode, but retain little actionable knowledge that reliably improves future behavior. This gap is especially problematic in stochastic, partially observable environments where policies must be discovered through repeated interactions rather than specified in advance.

Since fine-tuning is often infeasible and expensive, prompt-only self-improvement offers a promising alternative. However, three critical questions remain for applying this to stochastic, long-horizon sequential decision-making: (1) \textbf{What should be remembered?} Existing approaches typically commit to a single representation (heuristics/rules or behavior examples) without controlled comparison. (2) \textbf{How should memory propagate?} Unlike serial reflection, population-based training suggests parallel exploration could accelerate learning, but its prompt-only analogue is underexplored. (3) \textbf{Is the training method transferable across LLMs?} Whether gradient-free evolution yields consistent gains across diverse model families remains an open empirical question. 

We address these questions via FORGE, a \textbf{staged population protocol} where $N$ hierarchical ReAct agents evolve prompt-injected memory over $S$ stages. We evaluate four model families (Gemini-2.5-Flash-Lite, Grok-4-Fast, Llama-4-Maverick, Qwen3-235B) under three conditions: zero-shot, Reflexion \citep{shinn2023reflexion} (isolated single-stream learning), and FORGE (Reflexion plus population broadcast and graduation), comparing three memory representations across multiple independent sessions per model.

We study this challenge in CybORG CAGE-2 \citep{kiely2023autonomous}, a stochastic cyber-defense POMDP that combines four properties making it a demanding stress test for prompt-only adaptation: (1) a long horizon (30 steps) with partial observability, where LLM-based defense remains underexplored; (2) near-catastrophic zero-shot LLM performance, so memory-based learning is genuinely necessary; (3) scalar per-step reward with no natural-language feedback, requiring the agent to infer from numerical signals alone what went wrong; and (4) a practically important domain with a public leaderboard (DRL top score $-3.47$ \cite{kiely2023autonomous}) providing absolute reference points. In this setting, a ReAct agent based on four contemporary LLM families without any environment knowledge displays deeply negative zero-shot returns. All evidence in this paper is confined to CAGE-2 B\_line red agent at a 30-step horizon; generalization to other attacker types and environments remains future work.

\paragraph{Contributions.}
Our main contributions are: 
(1) \textbf{A population-based, gradient-free self-improvement protocol, FORGE}. The protocol evolves prompt-injected natural-language memory through staged learning with champion broadcast and graduation-based early stopping. We compare three conditions -- zero-shot, Reflexion (isolated reflection, no broadcast), and FORGE -- and observe 1.7--7.7$\times$ improvements in average returns over zero-shot, with the single best observed checkpoint return reaching $-3.60$ (against a maximum of $0$ and a DRL top score of $-3.47$); 
(2) \textbf{A controlled comparison of memory representations in a stochastic long-horizon environment.} \textsc{Rules}, \textsc{Examples}, and \textsc{Mixed} representations reach comparable final performance in the replicated Gemini study, with \textsc{Examples} achieving the best return ($-24.5$) and \textsc{Rules} offering the best cost-reliability profile with higher graduation rates and $\sim$40\% fewer tokens than \textsc{Examples};
(3) \textbf{Evidence that population-level transfer is critical.} Champion broadcast improves performance by 29--72\% over the Reflexion baseline in all 12 model-representation conditions and reduces catastrophic-failure rates to as low as $\sim$1\%; and (4) \textbf{Directional cross-family evidence.} FORGE improves over both zero-shot and Reflexion for all four tested model families, with disproportionately higher gains on weaker baselines.

\section{Related Work}
\paragraph{Prompt-Only Self-Improvement \& Baseline Selection.} A growing body of literature explores replacing weight updates with linguistic feedback. Reflexion \citep{shinn2023reflexion} stores critiques after failures, while Self-Refine \citep{madaan2023selfrefine} applies iterative critiques within a single response. CLIN \citep{majumder2024clin} extends prompt-only self-improvement to cross-episode causal-memory abstractions. Recent work on context evolution and test-time adaptation highlights that unconstrained self-edits can accumulate errors, motivating mechanisms that select and propagate only robust improvements across trials \citep{zhang2025ace,suzgun2025cheatsheet}. TextGrad \citep{yuksekgonul2024textgradautomaticdifferentiationtext} takes a complementary approach, performing gradient descent over text representations using LLM-generated feedback as the optimization signal. Voyager \citep{wang2023voyager} and ExpeL \citep{zhao2024expel} learn reusable skills or experience from successful episodes. Among these, Reflexion is the directly comparable baseline in our setting: it requires only a scalar success/failure indicator, applies failure-triggered verbal memory updates, and needs no task-specific engineering to operate with CAGE-2's per-step reward signal. The remaining methods would each require nontrivial adaptation: Voyager and ExpeL depend on reusable successes, scarce at CAGE-2 initialization where zero-shot performance is near-catastrophic; CLIN is built for structured text-simulator feedback rather than scalar per-step reward; Dynamic Cheatsheet \citep{suzgun2025cheatsheet} and ACE \citep{zhang2025ace} use update regimes that differ from scalar per-step cyber-defense reward; and TextGrad would require an auxiliary evaluator to convert environment reward into the textual optimization signal it expects (it is the most tractable future comparison). We therefore adopt Reflexion as the primary baseline and compare it against FORGE under identical model, memory representation, and training budget.

\paragraph{Memory Representations.} How to efficiently represent the knowledge that an agent should remember still remains an open question. Systems like AutoGuide \citep{fu2024autoguide}, ExpeL \citep{zhao2024expel}, and Voyager \citep{wang2023voyager} demonstrate the value of learning guidelines, experience, or skills represented by executable code. Complementary systems treat memory as a managed context resource or workflow artifact, e.g., MemGPT \citep{packer2024memgpt}, Agent Workflow Memory \citep{wang2025awm}, and Dynamic Cheatsheet \citep{suzgun2025cheatsheet}, but they rarely provide controlled comparisons of representation choices under identical training dynamics. While prior work compares instructions vs. exemplars in single-step tasks \citep{wan2024teachbetter} or even sequential decision-making \citep{sarukkai2025self}, we extend artifact efficiency comparison to adversarial POMDPs, evaluating \textsc{Rules}, \textsc{Examples}, and \textsc{Mixed} representations under identical staged learning conditions.

\paragraph{Population-Based Selection.} Population-Based Training (PBT) \citep{jaderberg2017pbt} is a canonical strategy for parallel exploration. Unlike classic hyperparameter PBT, our population mechanism selects among textual artifacts that shape the agent's policy via prompting. Whereas prompt-evolution methods typically optimize a single monolithic prompt and focus on single-step tasks \citep{fernando2024promptbreeder,guo2024evoprompt,yang2024opro}, we evolve structured lists of rules or examples, tied to specific failure modes, enabling targeted repairs without damaging unrelated competencies.

\paragraph{Cyber Defense Agents.} CybORG \citep{standen2021cyborg} and the CAGE-2 challenge \citep{kiely2023autonomous} provide a rigorous SOTA testbed currently dominated by reinforcement learning (RL) methods. Prior LLM-based cyber defenders are often evaluated either as fixed-prompt policies in CAGE-2 \citep{mohammadi2025leveraging} or in later CybORG variants (e.g., CAGE 4) as components within hybrid multi-agent systems \citep{castro2025llmcyber}. Our work demonstrates that LLM agents can improve their defense policies online without gradient updates.

\paragraph{Positioning.}
Our work combines three separate directions: (1) gradient-free self-improvement, (2) memory representation, and (3) population-based selection into a coherent protocol. Unlike memory-centric architectures that store complete records of agentic behavior \citep{park2023generative}, FORGE performs explicit trajectory analysis to update structured knowledge artifacts for specialized sub-agents, aligning with modular agent design patterns \citep{tran2025multiagent}. Our approach builds a Reflexion-inspired \citep{shinn2023reflexion} inner loop on ReAct \citep{yao2023react} scaffolding and wraps it with population-level selection and broadcast for hierarchical agents in an adversarial, stochastic cyber-defense POMDP.% ---- end paper_main/related_work ----

\begin{figure*}[t]
  \centering
  \includegraphics[width=1\textwidth]{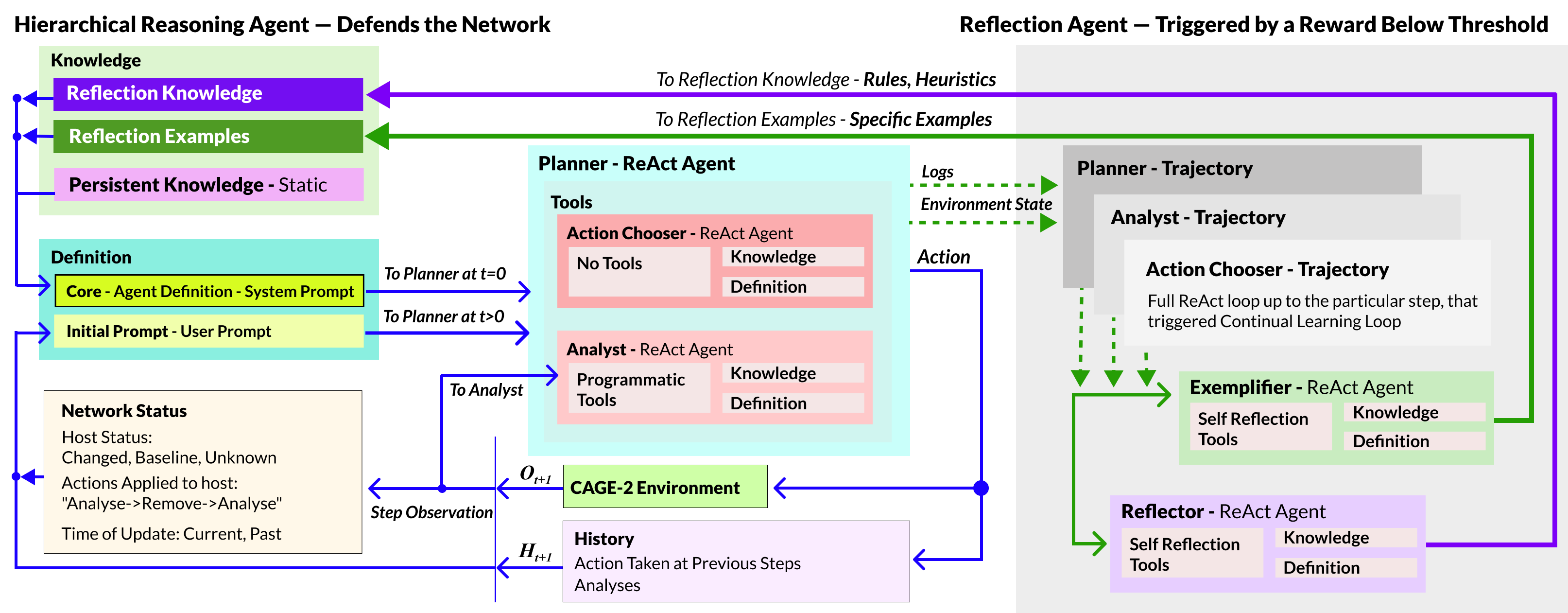}
  \caption{\textbf{System Overview.} (Left) Hierarchical ReAct agent with dynamic memory injection. (Right) Reflexion learning loop: upon a reward below threshold, a dedicated Reflector or Exemplifier agent analyzes the full trajectory and synthesizes knowledge artifacts that are injected back into the agent's memory.}
  \Description{Diagram showing the hierarchical ReAct agent architecture with Planner, Analyst, and ActionChooser sub-agents on the left, and the reflection and exemplification learning agents that convert failed trajectories into memory updates on the right.}
  \label{fig:system-overview}
\end{figure*}

\section{Method}
\begin{figure*}[h!]
  \centering
  \includegraphics[width=1\textwidth]{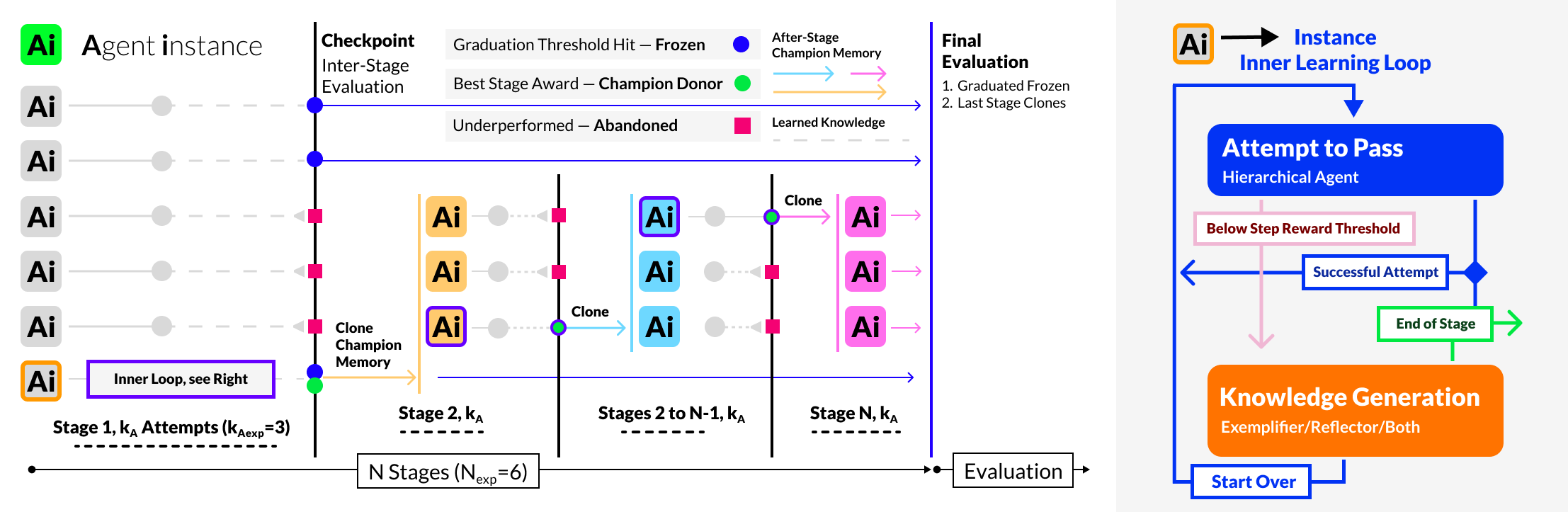}
  \caption{\textbf{Protocol Details.} (Left) The FORGE protocol involves parallel execution, champion selection, graduation and broadcast between stages. (Right) Inner learning loop inside each attempt.}
  \Description{Diagram showing the FORGE protocol with parallel agent instances executing across stages, champion selection and broadcast between stages on the left, and the inner abort-reflect-restart learning loop on the right.}
  \label{fig:protocol-overview}
\end{figure*}

We introduce FORGE (Failure-Optimized Reflective Graduation and Evolution), a multi-stage population protocol for evolving prompt-injected memory without weight updates. The method has three components: (1) a hierarchical ReAct agent with dynamic and persistent memory sections (Figure~\ref{fig:system-overview}, left); (2) an inner Reflexion-style learning loop \citep{shinn2023reflexion} that converts failures into reusable knowledge artifacts (Figure~\ref{fig:system-overview}, right); and (3) an outer population protocol that runs parallel instances in stages and propagates the best-discovered memory via champion broadcast (Figure~\ref{fig:protocol-overview}). A critical design constraint is that \textbf{no external oracle or stronger model is used}: the same LLM generates actions and synthesizes memory.

\subsection{Agent Architecture: Hierarchical ReAct with Dynamic Memory Injection}
\label{sec:agent-arch}
Figure \ref{fig:system-overview} (left) illustrates the hierarchical agent architecture. At each environment step, a top-level \textbf{Planner} selects the final defense action while delegating two sub-tasks to on-demand (implemented as tools) ReAct sub-agents: \textbf{Analyst} (interprets host-level observations) and \textbf{ActionChooser} (ranks valid actions with justification). All agents use the same underlying LLMs, and differ only in their role-specific system prompts and their injected memory.
\begin{algorithm}[h!]
  \small
  \caption{Failure-Triggered Reflexion Loop}
  \label{alg:reflexion-loop}
  \begin{algorithmic}[1]
    \Require Agent instance with memory $M_i$, attempts $k_A$, failure trigger $\tau$, representation $\in\{\textsc{Rules},\textsc{Examples},\textsc{Mixed}\}$
    \For{$a = 1$ to $k_A$}
      \State Track per-step reward $r_{step}$
      \State Run episode with memory $M_i$
      \If{$\exists\: step$ such that\ $r_{step}< \tau$}
        \State Abort; snapshot $\gets$ trajectories, $M_i$, metadata, environment state
        \State $\Delta \gets \textsc{UpdateMemory}(\text{snapshot}, \text{representation})$ \Comment{Reflector / Exemplifier}
        \State Apply edits: $M_i \gets \textsc{Apply}(M_i, \Delta)$
      \EndIf
    \EndFor
    \State \Return Updated memory $M_i$
  \end{algorithmic}
\end{algorithm}

The FORGE protocol runs $N$ copies of this agent hierarchy in parallel, each called an \emph{instance}. Instance $i \in \{ 1, ..., N\}$ maintains persistent and dynamic memory $M_i = (M_i^{P}, M_i^{A}, M_i^{C})$ for the Planner, Analyst, and ActionChooser. Persistent memory is instructions and specific knowledge set by the user; in this setup, only the ActionChooser and the learning agents (Reflector, Exemplifier) receive the environment action reference table, while the Planner has no pre-supplied action knowledge so that any strategic competence it acquires is attributable to learned artifacts. Dynamic memory is initially empty and accumulates knowledge artifacts generated by the learning agents during training. The \emph{representation} of these artifacts is the central experimental variable, taking one of three forms: (a) \textbf{\textsc{Rules}}: ordered lists of conditional heuristics; (b) \textbf{\textsc{Examples}}: structured demonstrations that mimic ReAct agent interaction (thought, tool, observation, answer) \citep{yao2023react}; or (c) \textbf{\textsc{Mixed}}: both rules and examples generated separately over the same context. Appendix~\ref{app:artifacts} provides verbatim examples of generated \textsc{Rules} and \textsc{Examples} artifacts.

Memory is stored on disk and re-injected into each agent's system prompt at every attempt, with a fixed capacity to prevent unbounded context growth. Sub-agents are instantiated on-demand within the Planner's reasoning loop with their own prompts and memory.

\subsection{Inner Loop: Failure-Triggered Reflexion}
\label{sec:inner-loop}
Figure~\ref{fig:system-overview} (right) illustrates the learning mechanism. Within each episode, the agent executes actions until completion or until a per-step reward $r_{step}$ drops below a failure threshold $\tau$. Upon failure, the episode is \textbf{aborted} immediately and the full trajectory is captured. A dedicated learning agent using the same underlying LLM, \textbf{Reflector} (for \textsc{Rules}) or \textbf{Exemplifier} (for \textsc{Examples}), analyzes the trajectory up to the failure point together with the environment state and synthesizes a knowledge artifact: either a conditional heuristic (\textsc{Rules}) or a structured interaction demonstration (\textsc{Examples}). In the \textsc{Mixed} condition, both agents generate artifacts over the same context. The generated artifact is appended to the agent's memory, and the episode restarts from step~0. This abort-and-restart cycle converts failures into a structured learning sequence, iterating up to $k_A$ attempts per stage. Algorithm~\ref{alg:reflexion-loop} formalizes this loop.

This loop constitutes a complete single-stream learning system implementing the Reflexion pattern \citep{shinn2023reflexion}. When run in isolation, each instance independently accumulates memory from its own trajectories without any cross-instance knowledge transfer. However, single-stream reflection in stochastic, partially observable environments lacks a selection pressure that distinguishes genuine policy improvement from noise: isolated instances can accumulate counterproductive artifacts that degrade performance below the zero-shot starting point, and even successful instances produce high-variance policies. A population-based protocol that runs multiple instances in parallel and propagates only the best-performing memory addresses this limitation structurally.

\subsection{Outer Loop: The FORGE Protocol}
\label{sec:forge-protocol}
To address the instability of isolated Reflexion, we propose FORGE, a \textbf{multi-stage, population-based protocol} that wraps the Reflexion inner loop (Algorithm~\ref{alg:reflexion-loop}) with three additional mechanisms: \emph{staged training} that creates periodic synchronization points, \emph{champion broadcast} that propagates the best-discovered memory to the population, and \emph{graduation} that freezes strong solutions and conserves compute. Figure~\ref{fig:protocol-overview} illustrates the protocol and Algorithm~\ref{alg:staged-population} formalizes it.

\begin{algorithm}[h!]
  \caption{FORGE: Staged Population Memory Training}
  \label{alg:staged-population}
  \begin{algorithmic}[1]
    \Require Instances $N$, stages $S$, attempts $k_A$, graduation threshold $\theta$, failure trigger $\tau$, representation, condition $\in\{\textsc{FORGE},\textsc{Reflexion}\}$
    \State Initialize dynamic memory $M_i \gets \emptyset$ for $i \in \{1,\ldots,N\}$
    \State Initialize graduated set $G \gets \emptyset$
    \For{$s = 1$ to $S$} \Comment{Outer loop: staged knowledge transfer}
      \State Initialize each instance $i$ in a uniquely seeded CAGE-2 environment with $M_i$
      \State Initialize attempt graduated set $H \gets \emptyset$
      \State Initialize checkpoint Return $R_{i}=0$
      \For{instance $i \notin G$ \textbf{in parallel}} \Comment{Concurrent Independent Instances}
        \State $M_i \gets \textsc{ReflexionLoop}(M_i, k_A, \tau, \text{representation})$ \Comment{Algorithm~\ref{alg:reflexion-loop}}
        \State $R_i \gets \textsc{Checkpoint}(M_i)$ \Comment{frozen evaluation, no learning}
      \EndFor
      \State $H \gets \{ i \notin G \;|\; R_i > \theta \}$ \Comment{new graduates}
      \State Freeze memories of $H$; $G \gets G \cup H$
      \If{condition = \textsc{FORGE} and $|\{i \notin G\}| > 0$}
        \State $i^* \gets \arg\max_{i \notin G} R_i$ \Comment{champion selection}
        \For{instance $i \notin G$}
          \State $M_i \gets M_{i^*}$ \Comment{broadcast: full memory replacement}
        \EndFor
      \EndIf
    \EndFor
    \State \Return Final evaluation of all $N$ frozen instances.
  \end{algorithmic}
\end{algorithm}

\begin{table*}[t]
  \centering
  \small
  \centering
  \caption{Experiment Count and Evaluated Episodes by Model and Condition}
  \label{tab:experiment_breakdown}
  \begin{tabular}{l c ccc ccc ccc}
  \toprule
  & \textbf{Zero-Shot} & \multicolumn{3}{c}{\textbf{Reflexion (instances)}} & \multicolumn{3}{c}{\textbf{FORGE (sessions)}} & \multicolumn{3}{c}{\textbf{FORGE w/o grad (sessions)}} \\
  \cmidrule(lr){3-5}\cmidrule(lr){6-8}\cmidrule(lr){9-11}
  \textbf{Model} & \textbf{Episodes}
    & \textbf{Rules} & \textbf{Ex} & \textbf{Mix}
    & \textbf{Rules} & \textbf{Ex} & \textbf{Mix}
    & \textbf{Rules} & \textbf{Ex} & \textbf{Mix} \\
  \midrule
  Gemini-2.5-Flash-Lite & 70 & 70 & 50 & 50 & 7 (70) & 7 (70) & 7 (140) & 2 (40) & 2 (40) & 2 (40) \\
  Grok-4-Fast & 100 & 30 & 30 & 30 & 3 (60) & 3 (60) & 3 (60) & 2 (40) & 2 (40) & 2 (40) \\
  Llama-4-Maverick & 50 & 70 & 30 & 30 & 3 (110) & 3 (60) & 3 (60) & 2 (40) & 2 (40) & 2 (40) \\
  Qwen3-235B & 50 & 30 & 30 & 50 & 4 (80) & 3 (60) & 3 (60) & 2 (40) & 2 (40) & 2 (40) \\
  \bottomrule
  \end{tabular}
 \end{table*}

FORGE adapts the Population-Based Training (PBT) framework \citep{jaderberg2017pbt} from weight space to prompt space. The PBT \emph{exploit} step maps to champion broadcast, which copies the best instance's memory artifacts to all active instances; the \emph{explore} step maps to the Reflexion inner loop, which independently evolves each instance's memory through failure-triggered reflection within the next stage. Two structural differences follow from operating on discrete textual artifacts rather than continuous weights: (1) broadcast performs full replacement rather than interpolation, because merging two natural-language rule sets would require a conflict-resolution mechanism that is itself unreliable, and (2) the explore step is not a random perturbation but a semantically grounded reflection on new failure trajectories.

\paragraph{Staged Training.}
Training is organized into $S$ sequential stages. At the beginning of each stage, every active instance is initialized in a uniquely seeded environment with its current memory $M_i$. Within the stage, each instance independently executes the Reflexion loop (Algorithm~\ref{alg:reflexion-loop}) for up to $k_A$ attempts. Stages serve as synchronization points: all instances complete their inner-loop attempts before any between-stage mechanism (checkpoint, graduation, broadcast) is applied.

\paragraph{Champion Broadcast.}
After each stage, a frozen checkpoint evaluation produces a return $R_i$ for each active instance. The instance with the highest checkpoint return among active (non-graduated) instances is designated the \emph{champion}, and its complete memory state replaces the memory of every other active instance. This is a destructive operation: each recipient discards its own accumulated artifacts and begins the next stage from the champion's memory. The design prioritizes convergence toward the best-discovered strategy over preserving population diversity.

\paragraph{Graduation and Early Stopping.}
Instances whose checkpoint return exceeds a graduation threshold $\theta$ are \textbf{graduated}: their memory is frozen and they are excluded from all subsequent stages. Graduation prevents regression by locking strong solutions before the next broadcast cycle can overwrite them, and conserves compute by excluding converged instances from further training.

When broadcast is disabled (condition = \textsc{Reflexion}), the protocol reduces to parallel independent Reflexion, the baseline against which FORGE is compared. After all stages complete, every instance (graduated or not) undergoes a final frozen evaluation.

\section{Experimental Setup}
\begin{table}[h!]
  \small
  \centering
  \caption{Models, FORGE Configuration, and Metrics}
  \label{tab:setup_summary}
  \begin{tabular}{p{0.18\linewidth} p{0.75\linewidth}}
  \toprule
  \textbf{Category} & \textbf{Configuration / Description} \\
  \midrule
  \textbf{Models} & Gemini-2.5-Flash-Lite (Primary), Grok-4-Fast, Llama-4-Maverick, Qwen3-235B \\
  \textbf{LLM Config} & \textbf{Agent (Planner/Analyst/ActionChooser)}: temp=0, max\_tokens=10000 \\
                      & \textbf{Learning (Reflector/Exemplifier)}: temp=0, max\_tokens=20000 \\
  \textbf{Conditions} & \textbf{Zero-shot}: Empty memory, no training \\
                      & \textbf{Reflexion}: Isolated single-stream reflection, no broadcast \\
                      & \textbf{FORGE}: Reflexion + champion broadcast + graduation \\
                      & \textbf{FORGE w/o grad}: Reflexion + champion broadcast, no graduation \\
  \textbf{Repre-sentations} & \textsc{Rules}, \textsc{Examples}, or \textsc{Mixed} memory \\
  \addlinespace
  \textbf{Protocol Params} & $N=10$ parallel instances, $S=6$ stages, $k_A=3$ attempts per stage \\
  \textbf{Hyperpara-meters} & Failure trigger $\tau = -1.1$ (per-step reward); Graduation $\theta = -15$ (episode return) \\
  \addlinespace
  \textbf{Metrics} & \textbf{Evaluation Return}: Mean return ($R=\sum r_{step}$, closer to 0 is better) \\
                   & \textbf{Graduation Rate}: \% of instances reaching $\theta=-15$ during checkpoint \\
                   & \textbf{Token Cost}: Total prompt + completion tokens (training + eval) \\
                   & \textbf{Tail Risk}: Frequency of catastrophic failures (return $< -100$) \\
  \bottomrule
  \end{tabular}
\end{table}
\paragraph{Evaluation Task and Environment: CybORG CAGE-2.}
We evaluate FORGE on CybORG CAGE-2 \citep{standen2021cyborg,kiely2023autonomous}, a cybersecurity gym environment modelled as POMDP where a blue defender protects a 13-host enterprise network against an automated red attacker over a 30-step horizon (canonical setting also includes 50- and 100-step episodes). CAGE-2 leaderboard is dominated by RL methods; public reference points against the \emph{B\_line} attacker over 30 steps include CardiffUni PPO $-3.47$ (DRL top score) \cite{cardiffuni_cage2_agent}, rule-based heuristic $-58.83$, random action $-154.06$, and no-action (sleeping) $-218.65$ \cite{kiely2023autonomous}. We define two failure severity levels anchored to this scale: return $<-100$ (major failure, between the heuristic and random baselines) and return $<-150$ (catastrophic failure, near-random performance). Conversely, returns above $-50$ represent significant improvement over the rule-based heuristic ($-58.83$), approaching the regime of competitive RL policies.

\begin{table*}[t]
  \centering
  \small
  \caption{Combined results: mean return $\pm$ SD by model, representation, and condition.
  \up\,= improvement, \dn\,= degradation over the reference column.
  \textbf{Bold} marks the best result per model across all FORGE variants.}
  \label{tab:forge-combined-results}
  \setlength{\tabcolsep}{3pt}
  \begin{tabular}{l r l rr rrr rrrr}
  \toprule
  & & & \multicolumn{2}{c}{\textbf{Reflexion}} & \multicolumn{3}{c}{\textbf{FORGE}} & \multicolumn{4}{c}{\textbf{FORGE w/o grad}} \\
  \cmidrule(lr){4-5}\cmidrule(lr){6-8}\cmidrule(lr){9-12}
  \textbf{Model} & \textbf{Zero-Shot} & \textbf{Representation}
    & \textbf{Mean$\pm$SD} & $\Delta$\textbf{ZS}
    & \textbf{Mean$\pm$SD} & $\Delta$\textbf{ZS} & $\Delta$\textbf{Refl}
    & \textbf{Mean$\pm$SD} & $\Delta$\textbf{ZS} & $\Delta$\textbf{Refl} & $\Delta$\textbf{FORGE} \\
  \midrule
  Gemini & \multirow{3}{*}{\fbox{$-189.6 \pm 53.9$}} & Rules & $-62.7 \pm 60.5$ & \textbf{\up67\%} & $-30.6 \pm 37.0$ & \textbf{\up84\%} & \textbf{\up51\%} & $-33.1 \pm 26.5$ & \textbf{\up83\%} & \textbf{\up47\%} & \dn8\% \\
   &  & Examples & $-78.9 \pm 60.4$ & \textbf{\up58\%} & $\mathbf{-24.5 \pm 21.1}$ & \textbf{\up87\%} & \textbf{\up69\%} & $-37.7 \pm 27.2$ & \textbf{\up80\%} & \textbf{\up52\%} & \dn54\% \\
   &  & Mixed & $-81.9 \pm 74.6$ & \textbf{\up57\%} & $-32.2 \pm 28.1$ & \textbf{\up83\%} & \textbf{\up61\%} & $-32.7 \pm 30.7$ & \textbf{\up83\%} & \textbf{\up60\%} & \dn1\% \\
  \midrule
  Grok & \multirow{3}{*}{\fbox{$-58.4 \pm 55.2$}} & Rules & $-79.9 \pm 62.6$ & \dn37\% & $-33.7 \pm 26.1$ & \textbf{\up42\%} & \textbf{\up58\%} & $-24.5 \pm 15.1$ & \textbf{\up58\%} & \textbf{\up69\%} & \textbf{\up27\%} \\
   &  & Examples & $-64.8 \pm 52.4$ & \dn11\% & $-42.7 \pm 43.3$ & \textbf{\up27\%} & \textbf{\up34\%} & $\mathbf{-14.0 \pm 2.0}$ & \textbf{\up76\%} & \textbf{\up78\%} & \textbf{\up67\%} \\
   &  & Mixed & $-114.4 \pm 76.8$ & \dn96\% & $-42.2 \pm 36.9$ & \textbf{\up28\%} & \textbf{\up63\%} & $-23.2 \pm 19.4$ & \textbf{\up60\%} & \textbf{\up80\%} & \textbf{\up45\%} \\
  \midrule
  Llama & \multirow{3}{*}{\fbox{$-113.1 \pm 81.5$}} & Rules & $-101.4 \pm 61.0$ & \textbf{\up10\%} & $-72.0 \pm 46.6$ & \textbf{\up36\%} & \textbf{\up29\%} & $-76.7 \pm 54.7$ & \textbf{\up32\%} & \textbf{\up24\%} & \dn7\% \\
   &  & Examples & $-53.9 \pm 59.7$ & \textbf{\up52\%} & $-28.3 \pm 15.7$ & \textbf{\up75\%} & \textbf{\up48\%} & $-42.2 \pm 35.3$ & \textbf{\up63\%} & \textbf{\up22\%} & \dn49\% \\
   &  & Mixed & $-44.2 \pm 40.6$ & \textbf{\up61\%} & $-29.6 \pm 25.6$ & \textbf{\up74\%} & \textbf{\up33\%} & $\mathbf{-23.8 \pm 9.3}$ & \textbf{\up79\%} & \textbf{\up46\%} & \textbf{\up19\%} \\
  \midrule
  Qwen & \multirow{3}{*}{\fbox{$-103.3 \pm 87.3$}} & Rules & $-88.4 \pm 83.5$ & \textbf{\up14\%} & $-25.2 \pm 21.0$ & \textbf{\up76\%} & \textbf{\up72\%} & $-17.3 \pm 2.6$ & \textbf{\up83\%} & \textbf{\up80\%} & \textbf{\up31\%} \\
   &  & Examples & $-57.6 \pm 69.5$ & \textbf{\up44\%} & $-24.3 \pm 35.8$ & \textbf{\up77\%} & \textbf{\up58\%} & $\mathbf{-15.5 \pm 2.3}$ & \textbf{\up85\%} & \textbf{\up73\%} & \textbf{\up36\%} \\
   &  & Mixed & $-80.4 \pm 89.4$ & \textbf{\up22\%} & $-29.3 \pm 20.4$ & \textbf{\up72\%} & \textbf{\up64\%} & $-17.9 \pm 2.8$ & \textbf{\up83\%} & \textbf{\up78\%} & \textbf{\up39\%} \\
  \bottomrule
  \end{tabular}
\end{table*}

\paragraph{Failure Trigger Sensitivity.}
The failure trigger $\tau=-1.1$ is derived from the environment's reward structure rather than tuned as a hyperparameter. Additionally, analysis of per-step penalties across 540 zero-shot episodes (Appendix~\ref{app:tau-analysis}) reveals that penalties fall into distinct groups: \emph{Restore} actions always cost $-1.0$ and are legitimate defensive operations, not failures; small failures cost $-1.1$ or $-1.2$; moderate failures $-2.0$ to $-3.2$; and severe failures $-11$ to $-14$ (with no values between $-3.3$ and $-10.9$). Since the trigger fires when $r_{step} < \tau$, setting $\tau=-1.1$ is the smallest threshold that excludes legitimate Restore penalties (exactly $-1.0$) while capturing real failures with 100\% precision (zero false positives) at 74\% recall (7,346 of 9,926 true triggers). A sensitivity sweep over $\tau \in \{-1.1, -2.0, -3.0, -11.0\}$, corresponding to the natural penalty groups, is reported alongside the main results. We set graduation threshold $\theta=-15$, which corresponds to roughly $10\times$ improvement over random agent performance \cite{kiely2023autonomous}.

\paragraph{Evaluation Modes.}
To explicitly distinguish selection from reporting, we define two evaluation modes: (1) \textbf{Checkpoint}: a frozen single-episode probe used during training to determine graduation and champion selection and (2) \textbf{Post-Session Evaluation}: a separate measurement of frozen instances, obtained after full FORGE session completion. Gemini-2.5-Flash-Lite serves as the primary study model with 7 independent FORGE sessions per representation followed by 1-2 evaluations per instance. Other models receive 3--4 FORGE sessions per representation as directional cross-family probes (Table~\ref{tab:experiment_breakdown}) followed by at least 2 evaluations per instance. The no-graduation ablation receives 2 sessions per model per representation across all four families followed by at least 2 evaluations per instance.

\paragraph{Baselines and Conditions.}
All results are compared against two baselines. The \emph{zero-shot baseline} evaluates the agent with empty memory and no training, measuring the total improvement attributable to memory evolution. The \emph{Reflexion baseline} runs the same failure-triggered reflection loop (Algorithm~\ref{alg:reflexion-loop}) but without champion broadcast: each instance evolves its own memory in isolation. Comparing FORGE against zero-shot quantifies the overall gain; comparing against Reflexion isolates the specific contribution of population-level knowledge transfer. We evaluate four LLM families under four conditions (Table~\ref{tab:setup_summary}): zero-shot, Reflexion, FORGE (Reflexion plus broadcast and graduation, Algorithm~\ref{alg:staged-population}), and FORGE without graduation (broadcast only). Each trained condition is crossed with three memory representations (\textsc{Rules}, \textsc{Examples}, \textsc{Mixed}). Table~\ref{tab:experiment_breakdown} breaks down session and episode counts per condition.

\begin{figure*}[t]
  \centering
  \includegraphics[width=1\textwidth]{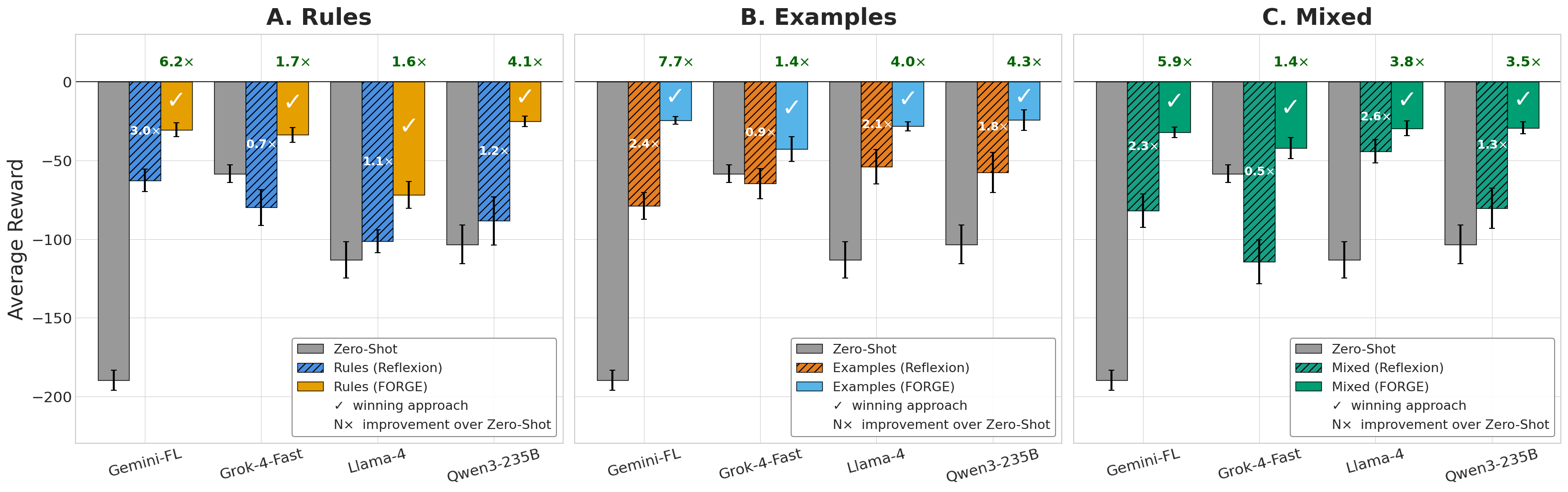}
  \caption{Comparison of memory representations (\textsc{Rules}, \textsc{Examples}, and \textsc{Mixed}) across zero-shot, Reflexion, and FORGE conditions for all four model families. Bars represent mean return; error bars denote SEM. Improvement factors over zero-shot annotated above FORGE bars; checkmarks indicate the winning condition.}
  \Description{Three-panel bar chart, one per representation, comparing mean return for zero-shot, Reflexion, and FORGE across four LLM families, with error bars, improvement factors, and winning-condition checkmarks.}
  \label{fig:strategy-panels}
\end{figure*}
\begin{figure*}[h!]
  \centering
  \includegraphics[width=\textwidth]{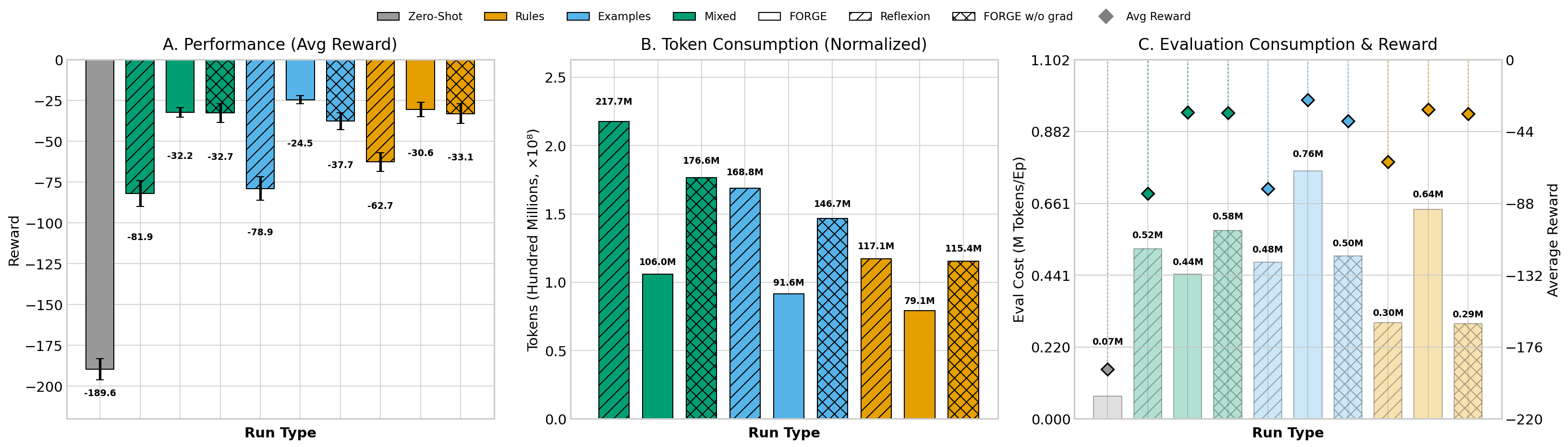}
  \caption{Combined analysis for Gemini-2.5-Flash-Lite. (A) Performance: All representations consistently outperform Baseline. (B) Token Cost: \textsc{Rules} representation is more efficient than others. (C) Evaluation Cost-Benefit: \textsc{Rules} offers best balance of low cost and high return.}
  \Description{Three-panel figure showing performance comparison, token cost breakdown, and cost-benefit scatter plot for Gemini-2.5-Flash-Lite across Rules, Examples, and Mixed representations.}
  \label{fig:combined-analysis-gemini}
\end{figure*}

\section{Results}
\label{sec:results}
We compare four conditions (zero-shot, Reflexion, FORGE, and FORGE without graduation) across four model families and three memory representations (Table~\ref{tab:forge-combined-results}). Results are organized around five questions: (1) the magnitude of improvement over zero-shot and Reflexion baselines across model families, (2) the comparative efficacy of memory representations (\textsc{Rules} vs. \textsc{Examples} vs. \textsc{Mixed}), (3) token cost and graduation dynamics, (4) cross-model generalization patterns, and (5) the contribution of population broadcast versus isolated Reflexion. We additionally report a no-graduation ablation that isolates the contribution of graduation from that of broadcast. A sensitivity sweep over the failure trigger threshold $\tau$ probes whether the chosen value ($-1.1$) is optimal or whether restricting reflection to more severe failures changes convergence behavior. We report \textbf{post-session evaluation} metrics unless explicitly referring to \textbf{checkpoint} probes used for intermediate champion selection.

\subsection{Main Findings}
\paragraph{Performance over Zero-Shot and Reflexion.} FORGE improves over both zero-shot and Reflexion baselines for every model family under all three representations (Table~\ref{tab:forge-combined-results}). Under the FORGE protocol, the strongest configurations reduce mean negative returns to the mid-twenties: Gemini improves from $-189.6$ to $-24.5$ (\textsc{Examples}, $7.7\times$), Qwen from $-103.3$ to $-24.3$ (\textsc{Examples}, $4.3\times$), Llama from $-113.1$ to $-28.3$ (\textsc{Examples}, $4.0\times$), and Grok from $-58.4$ to $-33.7$ (\textsc{Rules}, $1.7\times$). Compared to the Reflexion baseline, FORGE improves in all 12 model-representation conditions (Table~\ref{tab:forge-combined-results}, $\Delta$Refl columns). Figure~\ref{fig:strategy-panels} provides a detailed comparison across all models under each representation. The peak observed \textbf{checkpoint} return reaches $-3.60$ (Gemini \textsc{Rules}), approaching the DRL top score of $-3.47$, although post-session evaluation means remain lower due to sampling variance.

\paragraph{Representation Analysis.}
Figure~\ref{fig:strategy-panels} compares all three representations across all four models under zero-shot, Reflexion, and FORGE conditions. \textsc{Examples} achieves the best FORGE return for three of four models (Gemini, Llama, Qwen), while Grok performs best under \textsc{Rules}. In the replicated Gemini study (7 sessions per condition), all three representations yield large improvements over the zero-shot baseline ($-189.6$): \textsc{Examples} $-24.5 \pm 21.1$, \textsc{Rules} $-30.6 \pm 37.0$, and \textsc{Mixed} $-32.2 \pm 28.1$. \textsc{Examples} achieves the best mean return and lowest variance, while \textsc{Rules} provides the most reliable cost-performance profile with $\sim$40\% fewer total tokens than \textsc{Examples} (Figure~\ref{fig:combined-analysis-gemini}, Panel~B). This efficiency gap arises because example-based memory inflates prompt length, while rule-based artifacts are more compact and lead to faster graduation. \textsc{Mixed} falls between the two on both cost and performance (Figure~\ref{fig:combined-analysis-gemini}).

\paragraph{Token Cost Analysis.}
In the replicated Gemini study (Figure~\ref{fig:combined-analysis-gemini}, Panel~B), \textsc{Rules} consumes $\sim$106M total tokens compared to $\sim$177M for \textsc{Examples} and $\sim$188M for \textsc{Mixed}, a $\sim$40\% cost reduction. This efficiency stems from both shorter prompts and fewer active instances due to faster graduation. The no-graduation variants consume more tokens across all representations because all 10 instances remain active for all 6 stages, confirming that graduation's primary contribution is compute savings (Figure~\ref{fig:combined-analysis-gemini}, Panel~C).

\begin{figure*}[t]
  \centering
  \includegraphics[width=0.8\textwidth]{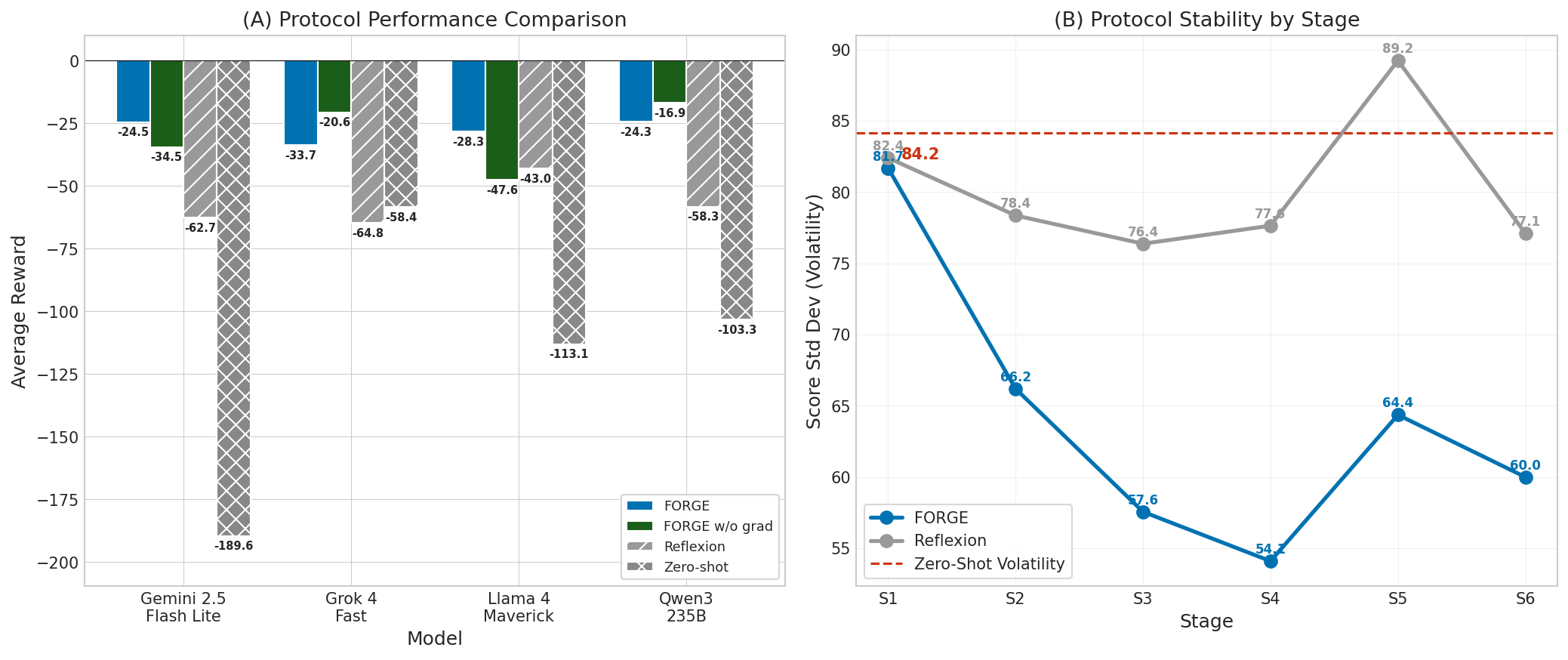}
  \caption{\textbf{Protocol comparison.} (A) Mean evaluation return across four models under four conditions: FORGE, FORGE without graduation, Reflexion (isolated learning), and zero-shot. FORGE and its no-graduation variant both outperform Reflexion and zero-shot for every model. (B) Standard deviation of checkpoint scores across stages; FORGE reduces volatility steadily while Reflexion remains near the zero-shot level (dashed).}
  \Description{Two-panel figure comparing FORGE, FORGE without graduation, Reflexion baseline, and zero-shot, showing mean returns and score volatility across models and stages.}
  \label{fig:protocol-panels}
\end{figure*}
\begin{figure*}[t]
  \centering
  \includegraphics[width=0.9\textwidth]{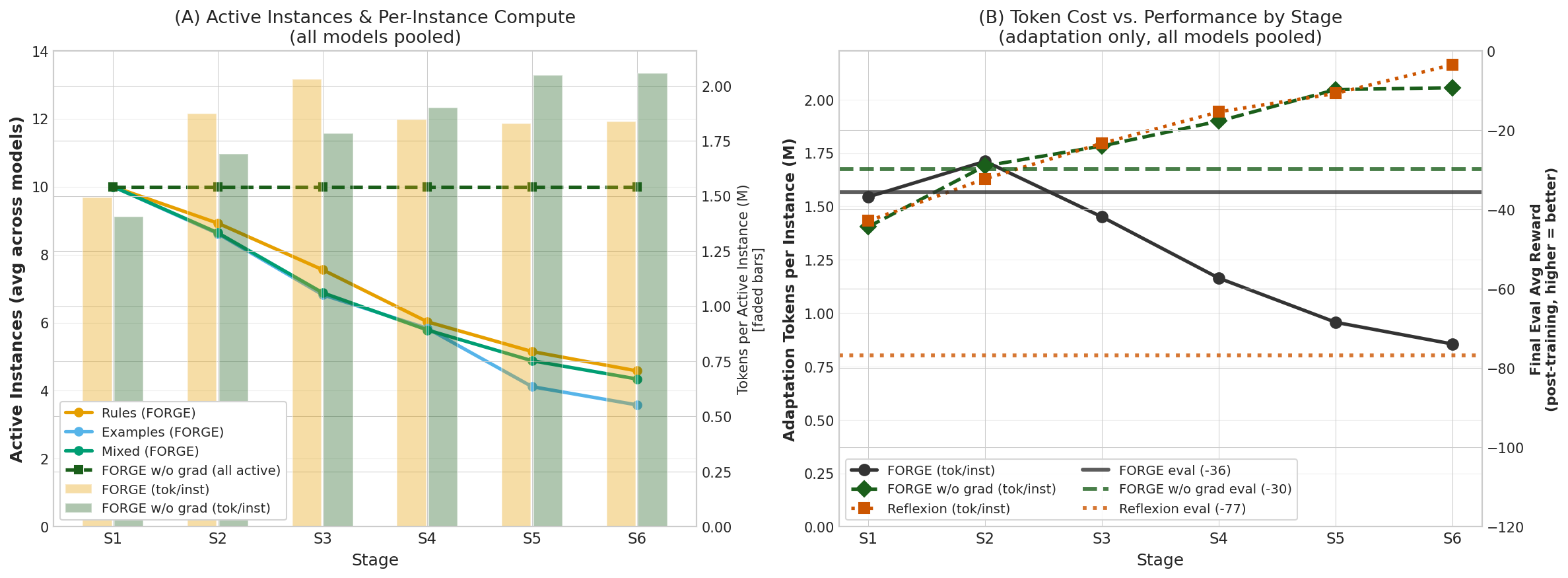}
  \caption{\textbf{Graduation dynamics and no-graduation ablation} (all models pooled). (A) Active instances and per-instance compute by stage: FORGE (solid lines) reduces active count via graduation, while the no-graduation variant (dashed) keeps all 10 instances active throughout. Faded bars show per-active-instance token cost. (B) Adaptation tokens per instance vs. final evaluation return by stage: FORGE's per-instance cost drops as instances graduate; the no-graduation variant achieves a slightly better pooled evaluation return at higher total cost. Reflexion baseline (dotted) shown for reference.}
  \Description{Two-panel figure comparing FORGE with and without graduation across stages, showing active instance counts, per-instance compute, adaptation token trajectories, and final evaluation returns.}
  \label{fig:graduation-combined}
  \end{figure*}

\paragraph{Cross-Model Analysis: Weak Models Benefit Most.}
We examine generalization across model families as directional evidence (non-Gemini models receive 3-4 FORGE sessions per representation). The magnitude of improvement inversely correlates with baseline strength: Gemini (worst baseline, $-189.6$) gains $7.7\times$, followed by Qwen ($4.3\times$) and Llama ($4.0\times$), while Grok (best baseline, $-58.4$) gains $1.7\times$ (Figure~\ref{fig:strategy-panels}). This pattern suggests FORGE functions primarily as a variance-reduction mechanism for unreliable policies, mitigating capability gaps rather than amplifying strong models. For every tested family, FORGE outperforms both zero-shot and Reflexion under all three representations.

\paragraph{Population Broadcast vs. Isolated Reflexion.}
Comparing FORGE against the Reflexion baseline isolates the contribution of population-level knowledge transfer (Figure~\ref{fig:protocol-panels}, Panel~A). FORGE improves post-session evaluation return by 29-72\% over Reflexion in all 12 model-representation conditions (Table~\ref{tab:forge-combined-results}, $\Delta$Refl columns). The no-graduation variant also outperforms Reflexion in all 12 conditions, confirming that champion broadcast is the essential mechanism. Reflexion instances exhibit persistently high volatility across stages (Figure~\ref{fig:protocol-panels}, Panel~B), whereas FORGE steadily compresses score variance.

\subsection{Ablation: No-Graduation}
The no-graduation variant retains champion broadcast but keeps all 10 instances active throughout all 6 stages, isolating the contribution of graduation from that of broadcast. Across models, FORGE reduces the active instance count as stages progress (Figure~\ref{fig:graduation-combined}, Panel~A, solid lines), with per-instance adaptation cost dropping as instances graduate. The no-graduation variant (dashed) consumes roughly twice the adaptation tokens per instance by S6 (Figure~\ref{fig:graduation-combined}, Panel~B). The effect of graduation on final performance is model-dependent: removing it helps Grok and Qwen (up to 67\% improvement over FORGE, Table~\ref{tab:forge-combined-results}), while Gemini and Llama perform better with graduation in 2 of 3 representations. This split suggests that graduation's memory-freezing mechanism protects strong early-stage artifacts in some models but terminates learning prematurely in others. Both FORGE variants outperform the Reflexion baseline in all 12 conditions, confirming that broadcast is the essential mechanism regardless of whether graduation is enabled.
\begin{figure*}[t]
  \centering
  \includegraphics[width=0.8\textwidth]{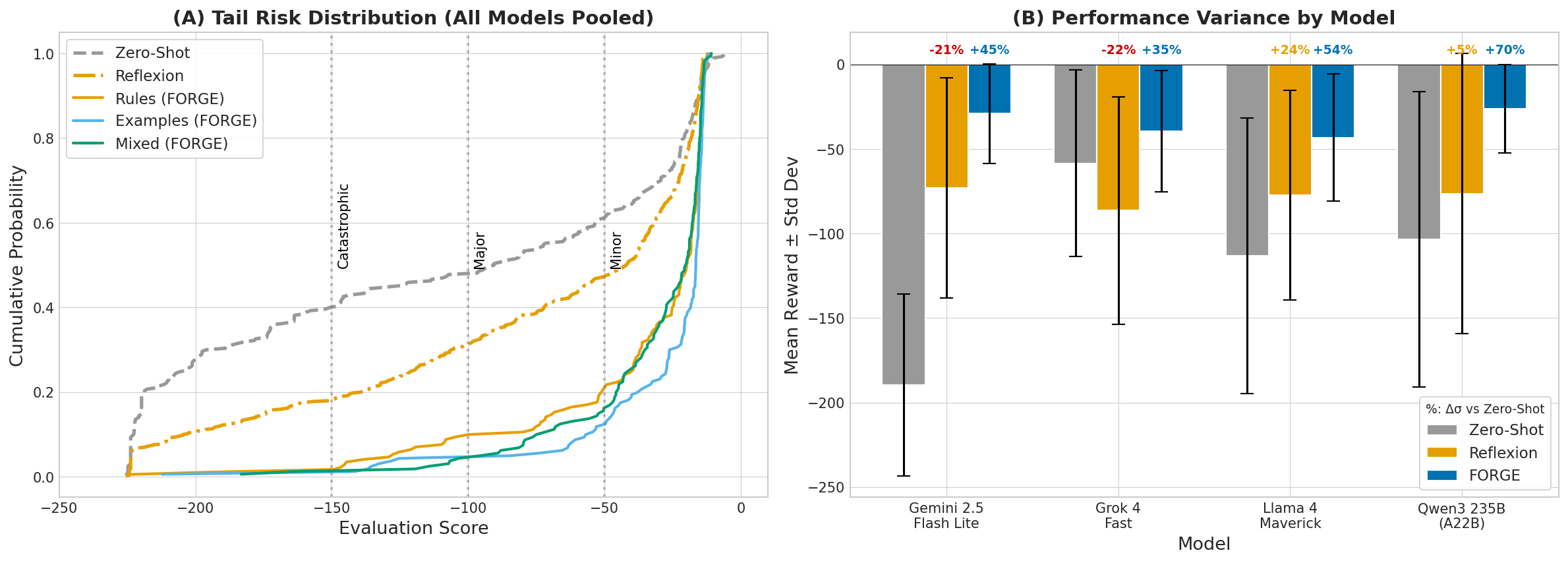}
  \caption{\textbf{Risk and variance analysis.} (A) Cumulative distribution of evaluation scores (all models pooled): zero-shot shows a heavy left tail; Reflexion partially compresses it; FORGE shifts the distribution sharply rightward. (B) Mean return $\pm$ SD by model under zero-shot, Reflexion, and FORGE (best representation per model).}
  \Description{Two-panel figure showing cumulative distribution of evaluation scores with heavy-tailed zero-shot, partially compressed Reflexion, and sharply rightward-shifted FORGE curves, and bar chart of mean returns with standard deviation across four models under three conditions.}
  \label{fig:risk-variance-analysis}
\end{figure*}

\subsection{Sensitivity: Failure Trigger Threshold}
The failure trigger $\tau$ determines which per-step penalties invoke reflection and restart the episode, directly controlling the learning signal's composition. To assess whether the chosen $\tau=-1.1$ is optimal, we sweep $\tau \in \{-1.1, -2.0, -3.0, -11.0\}$ on Gemini \textsc{Rules} (the primary configuration). The result is non-monotone: $\tau=-2.0$ (mean $-52.0$) and $\tau=-3.0$ (mean $-46.0$) both degrade relative to $\tau=-1.1$ (mean $-30.6$, 83\% graduation), while $\tau=-11.0$ yields the best result (mean $-24.6$, 93\% graduation). Restricting reflection to only the most severe failures appears to produce a cleaner learning signal, though skipping moderate failures hurts. The improvement at $\tau=-11.0$ suggests that harsher triggers and multi-threshold triggering strategies warrant further investigation. The reward distribution motivating $\tau=-1.1$ is detailed in Appendix~\ref{app:tau-analysis}.

\section{Discussion}
The protocol's efficacy stems from four interacting mechanisms: (1) reducing major failures by eliminating the heavy tail of low zero-shot returns (Figure~\ref{fig:risk-variance-analysis}, Panel~A); (2) compressing variance to stabilize outcomes (Figure~\ref{fig:risk-variance-analysis}, Panel~B); (3) population-level distillation via champion broadcast; and (4) graduation-based early stopping to reduce compute. Figure~\ref{fig:risk-variance-analysis}, Panel~A shows that Reflexion (orange) partially compresses the zero-shot tail, but FORGE shifts the distribution rightward, reducing the rate of episodes below $-100$ from $\sim$90\% (zero-shot) to $\sim$1\% under the strongest configurations.

The broadcast mechanism's consistent 29-72\% improvement over Reflexion across all 12 conditions suggests that the primary bottleneck in prompt-only adaptation is not the quality of individual reflections but the absence of a selection pressure that propagates rare discoveries to the population.

\section{Limitations \& Future Work}
Our study faces limitations in scope (single attacker type, fixed 30-step horizon in one application domain) and protocol dynamics (brittle single-best broadcast, checkpoint-evaluation misalignment). All evidence is confined to CAGE-2 B\_line; generalization to other attacker types and POMDP environments remains untested. Cross-family findings are presented as directional evidence based on 3-4 sessions per non-Gemini model. The failure trigger sensitivity sweep reveals that $\tau=-11.0$ outperforms the submitted $\tau=-1.1$, indicating that the chosen threshold is not optimal and that the broader design space of harsher triggers and multi-threshold triggering remains unexplored. Future work should address these limitations by testing additional attacker variants. Promising extensions include cross-strategy seeding (e.g., \textsc{Mixed} from \textsc{Rules}), cross-model artifact transfer, co-evolutionary adversarial training, cost-controlled comparisons against parameter-efficient fine-tuning to clarify the trade-offs of prompt-only adaptation, and replacing the Reflexion inner loop with alternative self-improvement methods (e.g., TextGrad \citep{yuksekgonul2024textgradautomaticdifferentiationtext}, Dynamic Cheatsheet \citep{suzgun2025cheatsheet}) to test whether the population broadcast mechanism generalizes further.

\section{Conclusion}
We introduced FORGE, a staged, population-based protocol for improving LLM agents via \emph{prompt-injected memory evolution}, demonstrating that effective long-horizon strategies can be learned without gradient updates or stronger teacher models. By coupling a Reflexion-style inner loop that converts failures into dynamic knowledge artifacts (\textsc{Rules}, \textsc{Examples}, or both) with an outer loop that stabilizes learning through champion broadcast and graduation-based early stopping, FORGE achieves 1.7-7.7$\times$ improvement over zero-shot baselines and 29-72\% improvement over the Reflexion baseline across all 12 model-representation conditions on CybORG CAGE-2 B\_line at a 30-step horizon. Among representations, \textsc{Examples} achieves the strongest returns for three of four models, while \textsc{Rules} offers the best cost-reliability profile with $\sim$40\% fewer tokens. The no-graduation ablation confirms that champion broadcast is the essential mechanism, with graduation primarily contributing compute savings. Cross-family results are directional evidence based on 3-4 sessions per non-Gemini model; generalization to other attacker types and environments remains future work. Within this benchmark, these results suggest that evolving interpretable natural-language memory provides a viable adaptation path for prompt-only learning in stochastic POMDPs where weight updates are infeasible, and encourage further research into alternative strategies. Reproducibility details and ethics considerations are in Appendix \ref{app:ethics}. The archived artifact is available at \url{https://doi.org/10.5281/zenodo.19907612}; the development repository is available at \url{https://github.com/isbogdanov/forge-protocol}.

\bibliographystyle{ACM-Reference-Format}
\bibliography{bib}

% ---- inlined from appendix/appendix ----
\appendix

\section{Ethics Statement \& Reproducibility}
\label{app:ethics}
All authors adhere to the ACM Code of Ethics\footnote{https://www.acm.org/code-of-ethics}. No human-subject data, personally identifiable information, or user-generated content is collected; all results are based on simulator-generated traces. Experiments are confined to the CybORG CAGE-2 benchmark and do not interact with real systems. We frame FORGE strictly for defensive decision-making but recognize the dual-use potential of cybersecurity automation. Since the protocol is compute-intensive, we report cost metrics and use graduation to reduce unnecessary runs. As non-native English speakers, we used LLM-based tools for language polishing and assistance with data processing scripts.

\subsection{Artifact Availability and Scope}

The artifact supporting this paper is archived on Zenodo at
\url{https://doi.org/10.5281/zenodo.19907612}. The development repository is
available at \url{https://github.com/isbogdanov/forge-protocol}. Detailed
build, configuration, and execution instructions are provided in the artifact
README.

The archived artifact contains the implementation of the FORGE protocol, the
experiment runner, container specification, API-key template, and configuration
files for running the population-broadcast and Reflexion-style baselines. \texttt{agent\_base/} contains the main implementation, including the hierarchical
Planner, Analyst, and ActionChooser agents, the Reflector and Exemplifier
learning agents, the CybORG coordinator, the learning coordinator, provider
configuration, and metric/logging utilities. The acting agents are configured
through YAML definition files, including static prompt components and dynamic
memory files e.g. \texttt{reflection\_knowledge.yaml} and
\texttt{reflection\_examples.yaml}. These files are updated during training when
the learning agents synthesize rules or examples from failed trajectories.

Experiments are launched through \texttt{run\_experiment.py}. The release
includes \texttt{experiment\_forge\_eval.yaml} for the FORGE condition
(\texttt{transfer\_strategy: best}, champion broadcast, 10 instances) and
\texttt{experiment\_reflexion\_eval.yaml} for the isolated Reflexion baseline
(\texttt{transfer\_strategy: individual}, no population broadcast). The
configuration files expose the main protocol parameters, including the number
of stages, the graduation threshold, the per-step failure threshold, the memory
representation (\texttt{rules}, \texttt{examples}, or \texttt{mixed}), and the
number of learning attempts per stage. The released FORGE configuration uses
six stages, a graduation threshold of $-15$, a per-step reflection trigger of
$-1.1$, and three learning attempts per stage.

A run creates an experiment directory containing the copied configuration,
stage-level summaries, workspaces with the learned memory snapshots, final
evaluation reports, aggregate summaries, runtime logs, and connector token-usage
logs for that run. Full raw LLM-provider transcripts and evolved memory
artifacts from the original study are not bundled in the public artifact because
of storage size and API cost. They may be available upon request from the
authors, subject to transfer and storage constraints. Because original run seeds
and provider-side execution state are not bundled, the artifact supports
executable reruns of the released protocol configurations rather than
bit-for-bit regeneration of the original logs. The paper itself reports the
per-instance evaluation scores used for the aggregate statistics; the artifact
is intended to let readers inspect the implementation, verify the protocol
configuration, and rerun selected FORGE or Reflexion conditions under the
documented setup.

% ===================
\section{CAGE-2 Environment Details}
\label{app:cage}
We evaluate the protocol on the CAGE-2 autonomous cyber-defense challenge~\citep{standen2021cyborg,cage2challenge}, a partially observable Markov decision process (POMDP) where a \emph{blue} defender protects an enterprise network against an automated \emph{red} attacker.
\begin{figure*}[t]
    \centering
    \begin{minipage}{0.8\textwidth}
        \centering
        \includegraphics[width=\textwidth]{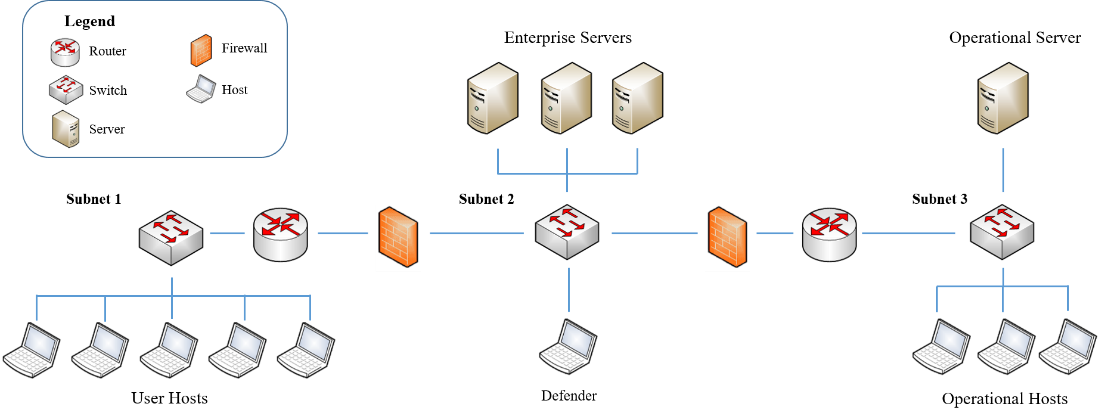}
        \caption*{(a) Network Topology: 3 subnets containing user hosts, enterprise servers, and operational servers.}
    \end{minipage}
    \begin{minipage}{0.8\textwidth}
        \centering
        \includegraphics[width=\textwidth]{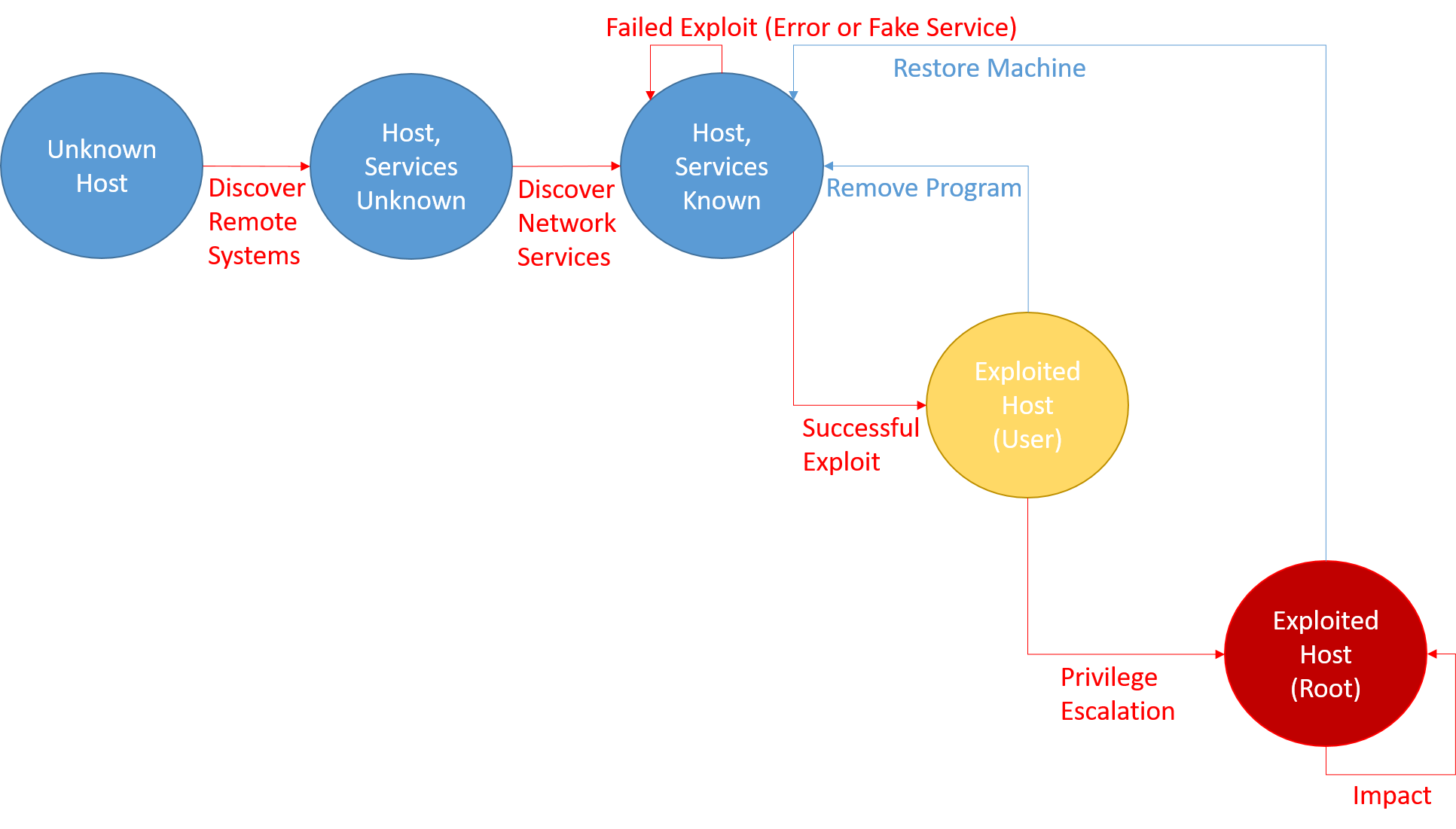}
        \caption*{(b) Attacker State Transitions: The progressive kill chain from discovery to root compromise.}
    \end{minipage}
    \caption{\textbf{CAGE-2 Environment Overview.} (a) The defender protects a 13-host network segmented into subnets \cite{kiely2023autonomous}. (b) The automated attacker follows a multi-stage state machine; successful exploits advance the attacker's position, while defender actions (like Restore) can reset this progress \cite{kiely2023autonomous}.}
    \Description{Two-part figure showing the CAGE-2 network topology with 13 hosts across 3 subnets, and the attacker state transition diagram from discovery through access, lateral movement, and privilege escalation.}
    \label{fig:cage-2-combined}
\end{figure*}

\paragraph{Topology and Assets.} The simulated network (Figure~\ref{fig:cage-2-combined}a) consists of 3 subnets hosting 13 nodes: user workstations (Subnet 1), enterprise servers (Subnet 2), and a high-value operational server (Subnet 3). The defender's goal is to maintain the availability and integrity of these assets, particularly the operational server.

\paragraph{Attacker Model (\texttt{b\_line}).} The red agent follows a scripted policy implementing a standard kill chain (Figure~\ref{fig:cage-2-combined}b). Starting from an external position, it progresses through phases of \emph{Discovery} (scanning for IPs), \emph{Access} (identifying services), \emph{Lateral Movement} (exploiting user privileges), and \emph{Escalation} (gaining root). The defender can push the attacker back to earlier states via remediation actions, creating a dynamic tug-of-war.

\paragraph{Action Space.} Episodes last $T{=}30$ steps. At each step, the blue agent selects an action:
\begin{itemize}
  \item \textbf{Monitor}: Passive observation with no intervention.
  \item \textbf{Analyse}~$h$: Inspect host~$h$ for anomalies (low cost).
  \item \textbf{Remove}~$h$: Terminate suspicious processes (moderate cost, imperfect success).
  \item \textbf{Restore}~$h$: Reimage~$h$ to a clean state (high cost, guarantees remediation).
  \item \textbf{Decoy}~$h$: Deploy a honeypot to misdirect the attacker (moderate cost).
\end{itemize}

\paragraph{Partial Observability and Reward.}
The defender observes only host-level indicators (processes, connections, files) and must infer the attacker's latent state. The per-step reward $r_{step} \le 0$ penalizes both ongoing compromise and the operational cost of defensive actions. An effective defense minimizes this penalty (achieving returns closer to zero) by balancing timely detection with targeted, cost-efficient interventions and proactive defense.

\section{Knowledge Artifact Examples}
\label{app:artifacts}
This section provides verbatim examples of the two artifact types generated during FORGE training, illustrating what is stored in prompt memory and how the representations differ. Both artifacts below were produced by a Qwen3-235B training session.

\paragraph{\textsc{Rules} Artifact.}
The Reflector distills each failed trajectory into conditional heuristics, organized per agent role. These rules are injected into the agent's system prompt via a \texttt{<reflection\_knowledge>} section.
\begin{figure}[h!]
\footnotesize
\begin{verbatim}
Planner rules:
- When lateral movement is confirmed,
  then Plan: Isolate path + Analyse
  adjacent hosts
- When reconnaissance is confirmed on an
  Enterprise host, then Plan: Deploy
  DecoySERVICE immediately
- When a host has undergone two consecutive
  Analyses and remains in changed state
  with high severity, then Plan: Execute
  Remove or Restore

Analyst rules:
- When Analyse confirms C2 beaconing,
  then Recommend Deploy DecoySERVICE
  on Server host
- When multiple critical anomalies are
  present, then set recommended_focus
  to 'containment'
\end{verbatim}
\caption{\textsc{Rules} artifact generated by the Reflector after a failed episode. Each rule is a conditional heuristic injected into the agent's system prompt.}
\Description{Verbatim listing of five conditional heuristic rules generated by the Reflector agent for the Planner and Analyst.}
\label{fig:artifact-rules}
\end{figure}

\paragraph{\textsc{Examples} Artifact.}
The Exemplifier converts the failed trajectory into a structured few-shot demonstration that mimics a complete ReAct interaction cycle (Thought--Tool--Observation--Answer), injected via a \texttt{<TOOL\_USE\_EXAMPLES>} section.
\begin{figure}[h!]
\footnotesize
\begin{verbatim}
<example description='PlanMonitorAndDecoy
  AfterReconAnalysis'>
  Thought: Enterprise_Host shows signs of
    reconnaissance from 10.0.247.46. Per
    reflection knowledge, plan monitoring
    and decoy deployment.
  Tool: get_suggestion_for_next_action:
    {"target_host": "Enterprise_Host",
     "situation": "recon confirmed",
     "severity": "high", ...}
  PAUSE
  Observation: [
    {"action": "Monitor",
     "confidence": 0.95},
    {"action": "DecoySSHD
       hostname=Enterprise_Host",
     "confidence": 0.90}, ...]
  Thought: Monitoring and decoy deployment
    are highest-priority.
  Answer: Monitor
</example>
\end{verbatim}
\caption{\textsc{Examples} artifact generated by the Exemplifier after a failed episode (abbreviated). The demonstration mimics a full ReAct interaction cycle.}
\Description{Verbatim listing of a structured few-shot demonstration artifact showing a complete ReAct interaction cycle.}
\label{fig:artifact-example}
\end{figure}

\section{Failure Trigger Threshold Analysis}
\label{app:tau-analysis}

\begin{figure*}[t]
  \centering
  \includegraphics[width=\textwidth]{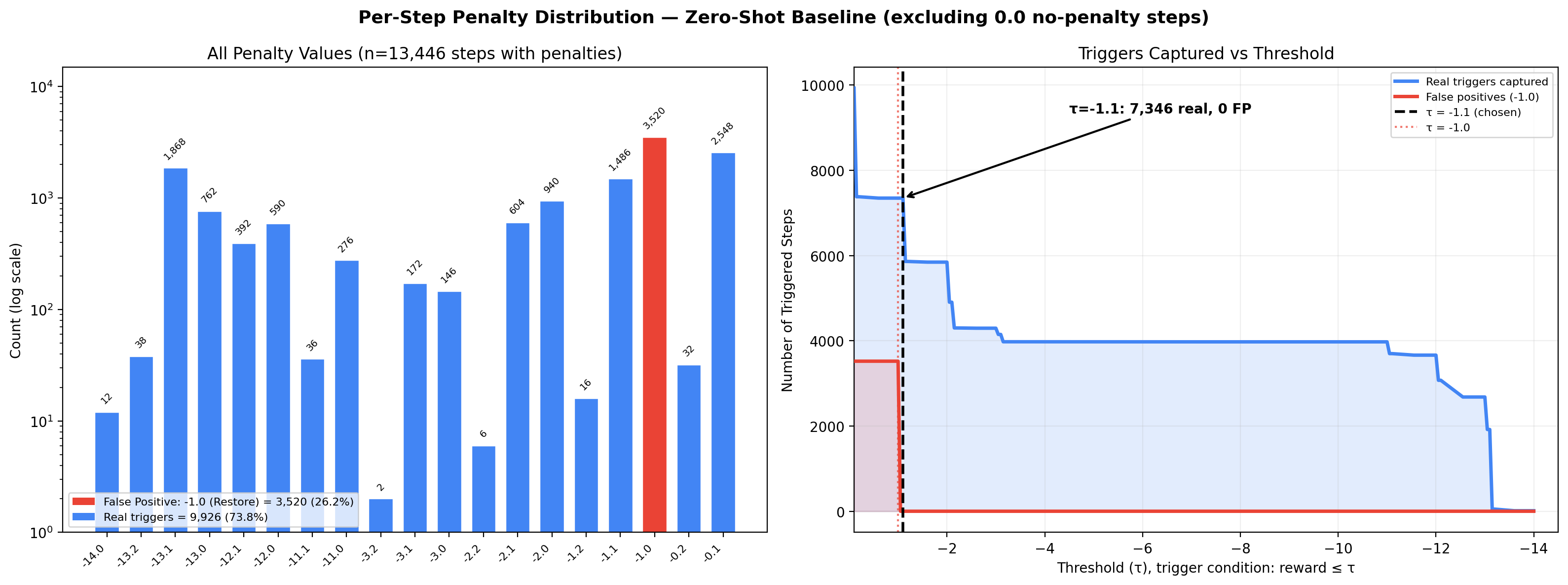}
  \caption{\textbf{Failure trigger threshold analysis.} (Left) Per-step penalty distribution across zero-shot episodes (log scale). The red bar at $-1.0$ represents legitimate Restore actions; blue bars are real failures. (Right) Triggers captured vs.\ threshold: $\tau=-1.1$ captures 7,346 real triggers with 0 false positives.}
  \Description{Two-panel figure showing the per-step penalty histogram on the left with Restore false positives highlighted in red, and a step function of triggers captured versus threshold on the right with the chosen tau=-1.1 annotated.}
  \label{fig:reward-histogram}
\end{figure*}

The failure trigger $\tau$ determines which per-step rewards abort the episode and invoke the reflection agent. Figure~\ref{fig:reward-histogram} shows the distribution of per-step penalties across 540 zero-shot episodes (all four model families, 13,446 penalized steps). Penalties cluster into distinct groups: \emph{Restore} actions at $-1.0$ (3,520 occurrences, red), small failures at $-1.1$ to $-1.2$, moderate failures at $-2.0$ to $-3.2$, and severe failures at $-11$ to $-14$, with a gap between $-3.3$ and $-10.9$. The chosen threshold $\tau=-1.1$ achieves 100\% precision (zero false positives from Restore) while capturing 74\% of all real failure events (7,346 of 9,926 true triggers).

To assess sensitivity, we tested three additional thresholds on Gemini \textsc{Rules} (the primary model-representation pair with 7-session coverage). Relative to $\tau=-1.1$ (mean return $-30.6$, 83\% graduation rate): $\tau=-2.0$ yields mean $-52.0$ over 40 episodes; $\tau=-3.0$ yields mean $-46.0$ over 40 episodes; and $\tau=-11.0$ yields mean $-24.6$ with 93\% graduation over 60 episodes. The result is non-monotone: skipping small and moderate failures ($\tau=-2.0$, $-3.0$) degrades performance, but restricting reflection to only the most severe failures ($\tau=-11.0$) improves over the submitted value. This suggests that triggering on severe failures produces a cleaner learning signal, though the broader design space of multi-threshold triggering remains unexplored.

Table~\ref{tab:tau-sweep-summary} provides session-level aggregates for each threshold, and Table~\ref{tab:tau-sweep-raw} reports the complete per-instance evaluation scores.

\begin{table}[htbp]
\centering
\scriptsize
\caption{Failure Trigger Threshold Sweep: Session-Level Results (Gemini Rules)}
\label{tab:tau-sweep-summary}
\begin{tabular}{lccccc}
\toprule
\textbf{Threshold ($\tau$)} & \textbf{Runs} & \textbf{Episodes} & \textbf{Mean Return} & \textbf{SD} & \textbf{Grad.\ Rate} \\
\midrule
$-1.1$ (default) & 7 & 140 & $-30.6$ & $37.0$ & 83\% \\
$-2.0$           & 2 & 40  & $-52.0$ & $59.6$ & 65\% \\
$-3.0$           & 2 & 40  & $-46.0$ & $52.9$ & 75\% \\
$-11.0$          & 3 & 60  & $\mathbf{-24.6}$ & $28.2$ & \textbf{93\%} \\
\bottomrule
\end{tabular}
\end{table}

\begin{table*}[htbp]
\centering
\scriptsize
\caption{Raw Evaluation Scores: Failure Trigger Threshold Sweep (Gemini Rules, FORGE Protocol)}
\label{tab:tau-sweep-raw}
\begin{tabular}{llcccccccc}
\toprule
\textbf{$\tau$} & \textbf{Instance} & \textbf{R1-E1} & \textbf{R1-E2} & \textbf{R2-E1} & \textbf{R2-E2} & \textbf{R3-E1} & \textbf{R3-E2} & \textbf{Tokens (M)} & \textbf{Grad} \\
\midrule
$-2.0$ & instance\_1 & $-35.1$ & $-13.2$ & $-51.2$ & $-14.7$ & — & — & \multirow{10}{*}{195.2} & \multirow{10}{*}{13/20} \\
       & instance\_2 & $-44.9$ & $-84.8$ & $-22.2$ & $-19.9$ & — & — & & \\
       & instance\_3 & $-14.8$ & $-16.6$ & $-20.4$ & $-37.0$ & — & — & & \\
       & instance\_4 & $-17.8$ & $-37.2$ & $-51.3$ & $-13.3$ & — & — & & \\
       & instance\_5 & $-17.8$ & $-172.2$ & $-198.8$ & $-16.9$ & — & — & & \\
       & instance\_6 & $-161.4$ & $-41.2$ & $-14.1$ & $-92.2$ & — & — & & \\
       & instance\_7 & $-19.3$ & $-17.5$ & $-16.7$ & $-14.2$ & — & — & & \\
       & instance\_8 & $-197.8$ & $-177.2$ & $-36.2$ & $-18.1$ & — & — & & \\
       & instance\_9 & $-14.2$ & $-17.3$ & $-21.1$ & $-15.2$ & — & — & & \\
       & instance\_10 & $-198.8$ & $-67.8$ & $-20.5$ & $-19.2$ & — & — & & \\
\addlinespace
$-3.0$ & instance\_1 & $-62.7$ & $-30.4$ & $-13.2$ & $-125.5$ & — & — & \multirow{10}{*}{226.7} & \multirow{10}{*}{15/20} \\
       & instance\_2 & $-36.1$ & $-172.2$ & $-77.4$ & $-14.5$ & — & — & & \\
       & instance\_3 & $-15.1$ & $-14.5$ & $-16.7$ & $-15.6$ & — & — & & \\
       & instance\_4 & $-12.3$ & $-111.2$ & $-17.5$ & $-17.7$ & — & — & & \\
       & instance\_5 & $-14.2$ & $-52.2$ & $-18.5$ & $-13.1$ & — & — & & \\
       & instance\_6 & $-22.3$ & $-223.6$ & $-16.4$ & $-47.1$ & — & — & & \\
       & instance\_7 & $-46.5$ & $-17.6$ & $-37.3$ & $-16.3$ & — & — & & \\
       & instance\_8 & $-15.5$ & $-13.9$ & $-14.7$ & $-16.0$ & — & — & & \\
       & instance\_9 & $-69.3$ & $-197.3$ & $-93.6$ & $-84.2$ & — & — & & \\
       & instance\_10 & $-15.3$ & $-14.5$ & $-11.2$ & $-16.2$ & — & — & & \\
\addlinespace
$-11.0$ & instance\_1 & $-72.5$ & $-14.2$ & $-12.5$ & $-26.1$ & $-27.2$ & $-38.2$ & \multirow{10}{*}{320.9} & \multirow{10}{*}{28/30} \\
        & instance\_2 & $-16.5$ & $-13.1$ & $-15.4$ & $-89.9$ & $-13.3$ & $-15.4$ & & \\
        & instance\_3 & $-15.6$ & $-59.7$ & $-13.4$ & $-13.2$ & $-17.5$ & $-15.9$ & & \\
        & instance\_4 & $-15.2$ & $-14.2$ & $-12.5$ & $-61.2$ & $-19.2$ & $-15.1$ & & \\
        & instance\_5 & $-12.7$ & $-14.2$ & $-13.5$ & $-21.7$ & $-17.4$ & $-16.7$ & & \\
        & instance\_6 & $-12.5$ & $-16.7$ & $-27.5$ & $-50.0$ & $-201.4$ & $-20.4$ & & \\
        & instance\_7 & $-16.2$ & $-15.4$ & $-14.4$ & $-17.5$ & $-49.5$ & $-14.0$ & & \\
        & instance\_8 & $-12.2$ & $-15.4$ & $-13.2$ & $-12.4$ & $-15.5$ & $-20.2$ & & \\
        & instance\_9 & $-13.4$ & $-12.4$ & $-14.2$ & $-13.4$ & $-11.8$ & $-16.2$ & & \\
        & instance\_10 & $-14.4$ & $-22.2$ & $-15.2$ & $-14.2$ & $-16.2$ & $-45.2$ & & \\
\bottomrule
\end{tabular}
\end{table*}

\section{Supplementary Analysis Tables}
\label{app:supplementary-tables}

Table~\ref{tab:forge-sd-results} reports standard deviations corresponding to the mean results in the main paper (Table~\ref{tab:forge-combined-results}). Table~\ref{tab:app-setup-summary} summarizes the experimental setup across all conditions.

\begin{table*}[htbp]
  \centering
  \scriptsize
  \caption{Standard deviations of episode return corresponding to
  Table~\ref{tab:forge-combined-results}. Lower SD = more reliable policy.
  \textbf{Bold} marks the lowest SD per model across all FORGE variants.
  \up/\dn\,= reduction/increase relative to the reference.}
  \label{tab:forge-sd-results}
  \begin{tabular}{l r l rr rrr rrrr}
  \toprule
  & & & \multicolumn{2}{c}{\textbf{Reflexion}} & \multicolumn{3}{c}{\textbf{FORGE}} & \multicolumn{4}{c}{\textbf{FORGE w/o grad}} \\
  \cmidrule(lr){4-5}\cmidrule(lr){6-8}\cmidrule(lr){9-12}
  \textbf{Model} & \textbf{ZS SD} & \textbf{Repr}
    & \textbf{SD} & $\Delta$\textbf{ZS}
    & \textbf{SD} & $\Delta$\textbf{ZS} & $\Delta$\textbf{Refl}
    & \textbf{SD} & $\Delta$\textbf{ZS} & $\Delta$\textbf{Refl} & $\Delta$\textbf{FORGE} \\
  \midrule
  Gemini & \multirow{3}{*}{53.9} & Rules & 60.5 & \dn12\% & 37.0 & \textbf{\up31\%} & \textbf{\up39\%} & 26.5 & \textbf{\up51\%} & \textbf{\up56\%} & \textbf{\up28\%} \\
   &  & Examples & 60.4 & \dn12\% & $\mathbf{21.1}$ & \textbf{\up61\%} & \textbf{\up65\%} & 27.2 & \textbf{\up50\%} & \textbf{\up55\%} & \dn29\% \\
   &  & Mixed & 74.6 & \dn38\% & 28.1 & \textbf{\up48\%} & \textbf{\up62\%} & 30.7 & \textbf{\up43\%} & \textbf{\up59\%} & \dn9\% \\
  \midrule
  Grok & \multirow{3}{*}{55.2} & Rules & 62.6 & \dn14\% & 26.1 & \textbf{\up53\%} & \textbf{\up58\%} & 15.1 & \textbf{\up73\%} & \textbf{\up76\%} & \textbf{\up42\%} \\
   &  & Examples & 52.4 & \textbf{\up5\%} & 43.3 & \textbf{\up21\%} & \textbf{\up17\%} & $\mathbf{2.0}$ & \textbf{\up96\%} & \textbf{\up96\%} & \textbf{\up96\%} \\
   &  & Mixed & 76.8 & \dn39\% & 36.9 & \textbf{\up33\%} & \textbf{\up52\%} & 19.4 & \textbf{\up65\%} & \textbf{\up75\%} & \textbf{\up47\%} \\
  \midrule
  Llama & \multirow{3}{*}{81.5} & Rules & 61.0 & \textbf{\up25\%} & 46.6 & \textbf{\up43\%} & \textbf{\up24\%} & 54.7 & \textbf{\up33\%} & \textbf{\up10\%} & \dn17\% \\
   &  & Examples & 59.7 & \textbf{\up27\%} & 15.7 & \textbf{\up81\%} & \textbf{\up74\%} & 35.3 & \textbf{\up57\%} & \textbf{\up41\%} & \dn125\% \\
   &  & Mixed & 40.6 & \textbf{\up50\%} & 25.6 & \textbf{\up69\%} & \textbf{\up37\%} & $\mathbf{9.3}$ & \textbf{\up89\%} & \textbf{\up77\%} & \textbf{\up64\%} \\
  \midrule
  Qwen & \multirow{3}{*}{87.3} & Rules & 83.5 & \textbf{\up4\%} & 21.0 & \textbf{\up76\%} & \textbf{\up75\%} & 2.6 & \textbf{\up97\%} & \textbf{\up97\%} & \textbf{\up88\%} \\
   &  & Examples & 69.5 & \textbf{\up20\%} & 35.8 & \textbf{\up59\%} & \textbf{\up48\%} & $\mathbf{2.3}$ & \textbf{\up97\%} & \textbf{\up97\%} & \textbf{\up94\%} \\
   &  & Mixed & 89.4 & \dn2\% & 20.4 & \textbf{\up77\%} & \textbf{\up77\%} & 2.8 & \textbf{\up97\%} & \textbf{\up97\%} & \textbf{\up86\%} \\
  \bottomrule
  \end{tabular}
\end{table*}

\begin{table}[htbp]
\centering
\scriptsize
\caption{Experimental Setup Summary}
\label{tab:app-setup-summary}
\begin{tabular}{p{0.25\linewidth} p{0.7\linewidth}}
\toprule
\textbf{Component} & \textbf{Description} \\
\midrule
\textbf{Models} & Gemini-2.5-Flash-Lite, Grok-4-Fast, Llama-4-Maverick, Qwen3-235B \\
\midrule
\textbf{Conditions} & \textbf{Zero-Shot}: No adaptation (empty memory) \\
& \textbf{Rules}: Contextual Instructions (Textual guidelines) \\
& \textbf{Examples}: Few-shot Demonstrations (State-Action pairs) \\
& \textbf{Mixed}: Combined Rules and Examples \\
\midrule
\textbf{Methods} & \textbf{FORGE}: Population-based continual learning (champion broadcast) \\
& \textbf{Reflexion}: Single-instance learning (no population broadcast) \\
\midrule
\textbf{Key Metrics} & \textbf{Evaluation Reward}: Cumulative reward per episode (Higher is better) \\
& \textbf{Graduation Rate}: \% of learning stages successfully completed \\
& \textbf{Token Efficiency}: Total tokens processed to reach performance \\
& \textbf{Volatility}: Standard deviation of rewards across stages \\
\midrule
\textbf{Scale} & \quad Zero-Shot: 5 experiments, 270 episodes \\
& \quad Reflexion: 38 experiments, 1,000 episodes \\
& \quad FORGE: 49 experiments, 890 episodes \\
& \quad FORGE w/o grad: 24 experiments, 480 episodes \\
& \textbf{Total}: 116 experiments, 2,640 episodes, 12364.0M tokens \\
\bottomrule
\end{tabular}
\end{table}

Table~\ref{tab:reflexion-detail} presents per-strategy Reflexion results across all models, providing detailed ablation data for the single-instance learning baseline.

\begin{table}[h!]
\centering
\scriptsize
\caption{Reflexion baseline evaluation detail. $N$\,= independent instances
evaluated (each runs the Reflexion loop in isolation; no cross-instance
interaction). Failure rate = episodes with return $<-100$.}
\label{tab:reflexion-detail}
\begin{tabular}{l l r r r c}
  \toprule
  \textbf{Model} & \textbf{Repr} & $N$ & \textbf{Mean} & \textbf{SD} & \textbf{Failure Rate ($<-100$)} \\
  \midrule
  Gemini & Rules & 70 & $-62.7$ & 60.5 & 21\% \\
   & Examples & 50 & $-78.9$ & 60.4 & 28\% \\
   & Mixed & 50 & $-81.9$ & 74.6 & 36\% \\
  \midrule
  Grok & Rules & 30 & $-79.9$ & 62.6 & 33\% \\
   & Examples & 30 & $-64.8$ & 52.4 & 27\% \\
   & Mixed & 30 & $-114.4$ & 76.8 & 50\% \\
  \midrule
  Llama & Rules & 70 & $-101.4$ & 61.0 & 50\% \\
   & Examples & 30 & $-53.9$ & 59.7 & 17\% \\
   & Mixed & 30 & $-44.2$ & 40.6 & 17\% \\
  \midrule
  Qwen & Rules & 30 & $-88.4$ & 83.5 & 33\% \\
   & Examples & 30 & $-57.6$ & 69.5 & 23\% \\
   & Mixed & 50 & $-80.4$ & 89.4 & 28\% \\
  \bottomrule
\end{tabular}
\end{table}

\subsection{Primary Model Statistical Detail}
\label{app:flash-lite-detail}

Table~\ref{tab:flash-lite-7run} provides session-level detail for the primary model (Gemini-2.5-Flash-Lite), which received the most comprehensive evaluation with 7 independent runs per strategy.

\begin{table}[htbp]
\centering
\scriptsize
\caption{Flash-Lite Session Results: Rules vs Examples vs Mixed (FORGE).
Avg = mean eval return across 10 instances in that session ($\pm$ within-session SD).}
\label{tab:flash-lite-7run}
\begin{tabular}{ll rrr}
\toprule
\textbf{Repr} & \textbf{Session} & \textbf{Avg Return} & \textbf{Grad} & \textbf{Tokens} \\
\midrule
Rules & 1 & -19.86$\pm$15.9 & 10/10 & 4.7M \\
 & 2 & -47.80$\pm$34.8 & 6/10 & 8.8M \\
 & 3 & -21.76$\pm$11.2 & 10/10 & 6.2M \\
 & 4 & -16.26$\pm$3.7 & 7/10 & 8.4M \\
 & 5 & -28.24$\pm$41.2 & 7/10 & 9.0M \\
 & 6 & -50.20$\pm$65.4 & 10/10 & 6.2M \\
 & 7 & -29.85$\pm$41.1 & 8/10 & 7.5M \\
\cmidrule(lr){2-5}
 & \textbf{Mean} & -30.57$\pm$13.4 & 8.3/10 & 7.3M$\pm$1.6M \\
\midrule
Examples & 1 & -28.73$\pm$20.4 & 7/10 & 8.3M \\
 & 2 & -18.72$\pm$13.7 & 9/10 & 8.6M \\
 & 3 & -24.24$\pm$15.0 & 10/10 & 6.0M \\
 & 4 & -23.18$\pm$18.7 & 9/10 & 6.7M \\
 & 5 & -36.11$\pm$38.1 & 8/10 & 9.3M \\
 & 6 & -24.45$\pm$21.8 & 8/10 & 10.5M \\
 & 7 & -16.01$\pm$4.1 & 10/10 & 9.4M \\
\cmidrule(lr){2-5}
 & \textbf{Mean} & -24.49$\pm$6.6 & 8.7/10 & 8.4M$\pm$1.6M \\
\midrule
Mixed & 1 & -20.04$\pm$13.3 & 8/10 & 11.4M \\
 & 2 & -38.67$\pm$18.4 & 7/10 & 12.6M \\
 & 3 & -42.06$\pm$52.9 & 9/10 & 8.3M \\
 & 4 & -34.37$\pm$16.4 & 6/10 & 11.4M \\
 & 5 & -38.46$\pm$36.8 & 7/10 & 8.4M \\
 & 6 & -27.43$\pm$18.0 & 7/10 & 10.1M \\
 & 7 & -24.36$\pm$18.5 & 10/10 & 8.8M \\
\cmidrule(lr){2-5}
 & \textbf{Mean} & -32.20$\pm$8.3 & 7.7/10 & 10.2M$\pm$1.7M \\
\bottomrule
\end{tabular}
\end{table}

\subsection{Graduation Dynamics}
\label{app:graduation}

Table~\ref{tab:graduation-distribution} reports the distribution of graduation rates across models and strategies, quantifying how many instances successfully completed the learning curriculum.

\begin{table}[htbp]
\centering
\scriptsize
\caption{Graduation Stage Distribution}
\label{tab:graduation-distribution}
\begin{tabular}{llccccccc}
\toprule
\textbf{Model} & \textbf{Representation} & \textbf{S1} & \textbf{S2} & \textbf{S3} & \textbf{S4} & \textbf{S5} & \textbf{S6} & \textbf{Never} \\
\midrule
Gemini & rules & 12 & 8 & 16 & 7 & 9 & 6 & 12 \\
Gemini & examples & 6 & 13 & 16 & 13 & 5 & 8 & 9 \\
Gemini & mixed & 10 & 12 & 12 & 9 & 8 & 3 & 16 \\
\midrule
Grok & rules & 3 & 8 & 8 & 3 & 2 & 2 & 4 \\
Grok & examples & 7 & 9 & 2 & 4 & 1 & 1 & 6 \\
Grok & mixed & 5 & 9 & 6 & 2 & 4 & 0 & 4 \\
\midrule
Llama & rules & 1 & 2 & 2 & 3 & 0 & 2 & 20 \\
Llama & examples & 1 & 3 & 1 & 5 & 3 & 2 & 15 \\
Llama & mixed & 4 & 4 & 1 & 4 & 0 & 4 & 13 \\
\midrule
Qwen3 & rules & 5 & 4 & 2 & 2 & 3 & 5 & 19 \\
Qwen3 & examples & 6 & 4 & 2 & 6 & 5 & 4 & 3 \\
Qwen3 & mixed & 3 & 3 & 1 & 1 & 1 & 0 & 21 \\
\bottomrule
\end{tabular}
\end{table}

\subsection{Computational Cost Breakdown}
\label{app:token-usage}

Table~\ref{tab:token-usage} breaks down token consumption by phase (Adaptation vs.\ Evaluation), providing transparency into the computational cost structure of the protocol.

\begin{table*}[htbp]
\centering
\scriptsize
\caption{Average Token Usage per Instance (Adaptation Phase, FORGE \& Reflexion)}
\label{tab:token-usage}
\begin{tabular}{lll cccc}
\toprule
\textbf{Model} & \textbf{Method} & \textbf{Representation} & \textbf{Avg Total/inst} & \textbf{Avg Prompt/inst} & \textbf{Avg Compl/inst} & \textbf{P/C Ratio} \\
\midrule
Gemini & FORGE & Rules & \textbf{7.3M$\pm$1.6M} & 6.4M & 0.8M & 7.8 \\
Gemini & Reflexion & Rules & 11.4M$\pm$0.5M & 10.1M & 1.3M & 7.9 \\
Gemini & FORGE & Examples & 8.4M$\pm$1.6M & 7.7M & 0.7M & 10.4 \\
Gemini & Reflexion & Examples & 13.1M$\pm$0.4M & 12.1M & 1.0M & 11.7 \\
Gemini & FORGE & Mixed & 10.2M$\pm$1.7M & 9.2M & 0.9M & 10.0 \\
Gemini & Reflexion & Mixed & 17.0M$\pm$0.5M & 15.8M & 1.2M & 13.2 \\
\midrule
Grok & FORGE & Rules & \textbf{4.7M$\pm$0.9M} & 3.8M & 0.9M & 4.2 \\
Grok & Reflexion & Rules & 8.8M$\pm$0.5M & 7.2M & 1.6M & 4.4 \\
Grok & FORGE & Examples & 8.4M$\pm$4.0M & 7.5M & 0.9M & 8.3 \\
Grok & Reflexion & Examples & 17.3M$\pm$0.9M & 15.5M & 1.8M & 8.8 \\
Grok & FORGE & Mixed & 9.3M$\pm$1.3M & 8.2M & 1.1M & 7.3 \\
Grok & Reflexion & Mixed & 19.9M$\pm$0.8M & 17.9M & 2.0M & 8.9 \\
\midrule
Llama & FORGE & Rules & 5.1M$\pm$0.2M & 4.8M & 0.3M & 17.2 \\
Llama & Reflexion & Rules & 5.5M$\pm$0.8M & 5.2M & 0.3M & 17.0 \\
Llama & FORGE & Examples & \textbf{4.6M$\pm$0.8M} & 4.4M & 0.2M & 23.1 \\
Llama & Reflexion & Examples & 5.7M$\pm$0.2M & 5.5M & 0.2M & 23.5 \\
Llama & FORGE & Mixed & 5.5M$\pm$0.9M & 5.3M & 0.2M & 23.8 \\
Llama & Reflexion & Mixed & 7.4M$\pm$0.0M & 7.1M & 0.3M & 25.7 \\
\midrule
Qwen3 & FORGE & Rules & \textbf{5.1M$\pm$1.2M} & 4.8M & 0.2M & 22.3 \\
Qwen3 & Reflexion & Rules & 5.8M$\pm$0.4M & 5.6M & 0.2M & 22.8 \\
Qwen3 & FORGE & Examples & 5.1M$\pm$1.6M & 4.9M & 0.2M & 23.0 \\
Qwen3 & Reflexion & Examples & 9.8M$\pm$0.5M & 9.4M & 0.3M & 27.3 \\
Qwen3 & FORGE & Mixed & 7.3M$\pm$3.8M & 7.1M & 0.2M & 32.6 \\
Qwen3 & Reflexion & Mixed & 9.6M$\pm$0.2M & 9.3M & 0.3M & 31.9 \\
\bottomrule
\end{tabular}
\end{table*}

\subsection{Baseline (Zero-Shot) Evaluation}
\label{app:baseline-scores}

Table~\ref{tab:raw-baseline} reports zero-shot evaluation scores for all models.

\begin{table}[htbp]
\centering
\caption{Raw Evaluation Scores: Zero-Shot (All Models)}
\label{tab:raw-baseline}
\scriptsize
\begin{tabular}{llc}
\toprule
\textbf{Model} & \textbf{Instance} & \textbf{S1} \\
\midrule
Gemini & instance\_1 & -215.10 \\
 & instance\_2 & -205.84 \\
 & instance\_3 & -171.80 \\
 & instance\_4 & -200.23 \\
 & instance\_5 & -198.80 \\
 & instance\_6 & -182.64 \\
 & instance\_7 & -174.09 \\
 & instance\_8 & -183.83 \\
 & instance\_9 & -171.19 \\
 & instance\_10 & -192.60 \\
\bottomrule
\end{tabular}

\begin{tabular}{llcc}
\toprule
\textbf{Model} & \textbf{Instance} & \textbf{S1} & \textbf{S2} \\
\midrule
Grok & instance\_1 & -93.92 & -31.84 \\
 & instance\_2 & -62.18 & -64.32 \\
 & instance\_3 & -83.82 & -86.48 \\
 & instance\_4 & -51.84 & -28.00 \\
 & instance\_5 & -33.50 & -16.46 \\
 & instance\_6 & -58.64 & -51.70 \\
 & instance\_7 & -89.34 & -116.58 \\
 & instance\_8 & -77.94 & -35.76 \\
 & instance\_9 & -52.30 & -43.00 \\
 & instance\_10 & -57.58 & -33.66 \\
\bottomrule
\end{tabular}

\begin{tabular}{llc}
\toprule
\textbf{Model} & \textbf{Instance} & \textbf{S1} \\
\midrule
Llama & instance\_1 & -113.34 \\
 & instance\_2 & -111.98 \\
 & instance\_3 & -92.24 \\
 & instance\_4 & -184.46 \\
 & instance\_5 & -122.64 \\
 & instance\_6 & -140.82 \\
 & instance\_7 & -113.66 \\
 & instance\_8 & -104.24 \\
 & instance\_9 & -60.48 \\
 & instance\_10 & -87.24 \\
\bottomrule
\end{tabular}

\begin{tabular}{llc}
\toprule
\textbf{Model} & \textbf{Instance} & \textbf{S1} \\
\midrule
Qwen3 & instance\_1 & -88.48 \\
 & instance\_2 & -97.98 \\
 & instance\_3 & -94.80 \\
 & instance\_4 & -103.26 \\
 & instance\_5 & -60.58 \\
 & instance\_6 & -73.00 \\
 & instance\_7 & -152.06 \\
 & instance\_8 & -83.60 \\
 & instance\_9 & -180.50 \\
 & instance\_10 & -99.16 \\
\bottomrule
\end{tabular}
\end{table}

\begin{table}[htbp]
  \centering
  \scriptsize
  \caption{Complete Experimental Results: Llama}
  \label{tab:full-results-llama}
  \begin{tabular}{clllcc}
  \toprule
  \textbf{\#} & \textbf{Method} & \textbf{Representation} & \textbf{Avg Reward} & \textbf{Tokens} & \textbf{Graduated} \\
  \midrule
  1 & FORGE & Examples & -33.12 & 5.5M & 2/10 \\
  2 & FORGE & Examples & -19.16 & 4.3M & 8/10 \\
  3 & FORGE & Examples & -32.62 & 4.0M & 5/10 \\
  4 & FORGE & Mixed & -27.82 & 5.4M & 5/10 \\
  5 & FORGE & Mixed & -36.96 & 4.7M & 7/10 \\
  6 & FORGE & Mixed & -23.91 & 6.5M & 5/10 \\
  7 & FORGE & Rules & -92.59 & 5.1M & 4/10 \\
  8 & FORGE & Rules & -82.50 & 5.3M & 1/10 \\
  9 & FORGE & Rules & -40.78 & 5.0M & 5/10 \\
  \midrule
  10 & Reflexion & Examples & -48.07 & 5.8M & --- \\
  11 & Reflexion & Examples & -65.69 & 5.5M & --- \\
  12 & Reflexion & Mixed & -46.81 & 7.4M & --- \\
  13 & Reflexion & Mixed & -39.11 & 7.4M & --- \\
  14 & Reflexion & Rules & -63.32 & 7.0M & --- \\
  15 & Reflexion & Rules & -113.69 & 5.3M & --- \\
  16 & Reflexion & Rules & -88.19 & 5.3M & --- \\
  17 & Reflexion & Rules & -121.42 & 5.2M & --- \\
  \midrule
  18 & Zero-Shot & --- & -113.11 & 0.4M & --- \\
  \bottomrule
  \end{tabular}
  \end{table}
  
  \begin{table}[htbp]
  \centering
  \scriptsize
  \caption{Complete Experimental Results: Qwen3}
  \label{tab:full-results-qwen3}
  \begin{tabular}{clllcc}
  \toprule
  \textbf{\#} & \textbf{Method} & \textbf{Representation} & \textbf{Avg Reward} & \textbf{Tokens} & \textbf{Graduated} \\
  \midrule
  1 & FORGE & Examples & -17.04 & 6.1M & 9/10 \\
  2 & FORGE & Examples & -20.22 & 5.8M & 9/10 \\
  3 & FORGE & Examples & -35.58 & 3.3M & 9/10 \\
  4 & FORGE & Mixed & -47.82 & 11.6M & 2/10 \\
  5 & FORGE & Mixed & -19.41 & 5.9M & 3/10 \\
  6 & FORGE & Mixed & -20.70 & 4.3M & 4/10 \\
  7 & FORGE & Rules & -34.67 & 6.6M & 2/10 \\
  8 & FORGE & Rules & -26.99 & 4.9M & 2/10 \\
  9 & FORGE & Rules & -21.62 & 5.1M & 9/10 \\
  10 & FORGE & Rules & -17.35 & 3.6M & 8/10 \\
  \midrule
  11 & Reflexion & Examples & -56.15 & 10.0M & --- \\
  12 & Reflexion & Examples & -60.41 & 9.3M & --- \\
  13 & Reflexion & Mixed & -69.15 & 9.5M & --- \\
  14 & Reflexion & Mixed & -125.41 & 9.9M & --- \\
  15 & Reflexion & Rules & -86.19 & 6.1M & --- \\
  16 & Reflexion & Rules & -89.48 & 5.6M & --- \\
  \midrule
  17 & Zero-Shot & --- & -103.34 & 0.4M & --- \\
  \bottomrule
  \end{tabular}
  \end{table}

  \section{Raw Evaluation Scores}
\label{app:raw-scores}

Tables~\ref{tab:raw-gemini-best}--\ref{tab:raw-others-best} report post-session evaluation scores for trained agents under the \textsc{Best} transfer protocol.

This appendix provides complete per-instance, per-run evaluation scores for all 2,640 evaluated episodes across 116 experiments. These raw data support the aggregate statistics reported in Table~\ref{tab:forge-combined-results} and enable full transparency regarding variances, outliers, and tail behavior. Each table reports episode returns ($R = \sum_{t=1}^{30} r_t$) with learning disabled (frozen evaluation). Column headers E1, E2, etc., denote independent evaluation runs; "—" indicates no data for that run.

\begin{table}[htbp]
\centering
\scriptsize
\caption{Raw Evaluation Scores: Gemini Flash-Lite, FORGE}
\label{tab:raw-gemini-best}
\begin{tabular}{llccccccc}
\toprule
\textbf{Repr} & \textbf{Instance} & \textbf{S1} & \textbf{S2} & \textbf{S3} & \textbf{S4} & \textbf{S5} & \textbf{S6} & \textbf{S7} \\
\midrule
Rules & instance\_1 & -64.90 & -73.30 & -18.20 & -14.60 & -18.10 & -69.50 & -17.50 \\
 & instance\_2 & -16.20 & -18.20 & -15.20 & -13.20 & -14.20 & -25.20 & -13.20 \\
 & instance\_3 & -13.20 & -39.40 & -13.30 & -17.20 & -13.20 & -13.30 & -15.20 \\
 & instance\_4 & -15.30 & -108.20 & -36.20 & -25.40 & -14.20 & -14.50 & -22.00 \\
 & instance\_5 & -12.40 & -100.00 & -12.50 & -16.20 & -12.20 & -14.10 & -17.70 \\
 & instance\_6 & -15.60 & -37.20 & -18.20 & -18.20 & -14.10 & -15.50 & -15.70 \\
 & instance\_7 & -14.20 & -49.80 & -43.20 & -13.70 & -20.80 & -12.40 & -146.50 \\
 & instance\_8 & -16.20 & -15.50 & -14.70 & -15.20 & -15.60 & -38.30 & -14.30 \\
 & instance\_9 & -16.60 & -22.10 & -32.70 & -16.50 & -145.30 & -223.80 & -17.20 \\
 & instance\_10 & -14.00 & -14.30 & -13.40 & -12.40 & -14.70 & -75.40 & -19.20 \\
\midrule
Examples & instance\_1 & -15.20 & -13.40 & -46.80 & -14.30 & -36.60 & -27.40 & -26.20 \\
 & instance\_2 & -60.80 & -13.20 & -16.50 & -14.20 & -14.40 & -16.60 & -15.40 \\
 & instance\_3 & -15.60 & -17.30 & -14.00 & -63.50 & -14.60 & -27.50 & -17.30 \\
 & instance\_4 & -15.50 & -15.40 & -40.20 & -14.20 & -135.50 & -12.20 & -15.20 \\
 & instance\_5 & -13.90 & -57.40 & -49.30 & -14.40 & -13.90 & -18.30 & -13.30 \\
 & instance\_6 & -13.70 & -13.40 & -15.60 & -14.80 & -15.40 & -13.60 & -14.30 \\
 & instance\_7 & -63.10 & -15.60 & -21.40 & -53.20 & -14.20 & -13.50 & -15.20 \\
 & instance\_8 & -14.90 & -16.50 & -13.40 & -15.20 & -53.00 & -17.20 & -11.40 \\
 & instance\_9 & -27.10 & -13.10 & -12.40 & -13.40 & -15.20 & -13.60 & -13.50 \\
 & instance\_10 & -47.50 & -11.90 & -12.80 & -14.60 & -48.30 & -84.60 & -18.30 \\
\midrule
Mixed & instance\_1 & -17.85 & -31.80 & -27.40 & -16.20 & -10.70 & -68.00 & -23.70 \\
 & instance\_2 & -14.65 & -79.70 & -76.35 & -48.25 & -13.70 & -15.30 & -15.05 \\
 & instance\_3 & -14.20 & -19.25 & -17.05 & -42.95 & -14.25 & -17.50 & -17.90 \\
 & instance\_4 & -55.70 & -45.75 & -14.70 & -16.35 & -14.85 & -13.50 & -50.25 \\
 & instance\_5 & -14.25 & -30.00 & -28.40 & -35.75 & -15.45 & -16.40 & -14.35 \\
 & instance\_6 & -13.45 & -52.85 & -25.45 & -50.30 & -88.05 & -45.80 & -66.20 \\
 & instance\_7 & -27.95 & -44.75 & -16.75 & -15.90 & -30.65 & -11.00 & -13.90 \\
 & instance\_8 & -15.05 & -17.75 & -15.45 & -21.85 & -107.20 & -36.55 & -13.30 \\
 & instance\_9 & -13.50 & -27.20 & -15.75 & -34.25 & -14.75 & -21.70 & -14.75 \\
 & instance\_10 & -13.80 & -37.70 & -183.25 & -61.85 & -75.00 & -28.55 & -14.25 \\
\bottomrule
\end{tabular}
\end{table}
  % Table A4: Raw Eval - Best Transfer (Grok/Llama/Qwen)
\begin{table*}[htbp]
\centering
\caption{Raw Evaluation Scores: FORGE (Grok, Llama, Qwen)}
\label{tab:raw-others-best}
\scriptsize
\begin{tabular}{ll cccc ccc ccc}
\toprule
\textbf{Model} & \textbf{Instance} & \multicolumn{4}{c}{\textbf{Rules}} & \multicolumn{3}{c}{\textbf{Examples}} & \multicolumn{3}{c}{\textbf{Mixed}} \\
\cmidrule(lr){3-6}\cmidrule(lr){7-9}\cmidrule(lr){10-12}
 & & \textbf{S1} & \textbf{S2} & \textbf{S3} & \textbf{S4} & \textbf{S1} & \textbf{S2} & \textbf{S3} & \textbf{S1} & \textbf{S2} & \textbf{S3} \\
\midrule
Grok & instance\_1 & -39.10 & -45.60 & -14.00 & --- & -125.35 & -16.55 & -14.55 & -14.70 & -18.10 & -29.80 \\
 & instance\_2 & -14.30 & -28.00 & -39.40 & --- & -44.80 & -17.85 & -14.30 & -81.40 & -79.85 & -13.55 \\
 & instance\_3 & -144.65 & -39.35 & -32.85 & --- & -141.80 & -46.35 & -15.05 & -33.40 & -114.50 & -18.00 \\
 & instance\_4 & -32.00 & -14.95 & -43.40 & --- & -132.70 & -26.15 & -37.75 & -21.70 & -43.20 & -16.30 \\
 & instance\_5 & -34.00 & -13.65 & -13.60 & --- & -46.15 & -41.30 & -13.80 & -15.60 & -71.60 & -14.70 \\
 & instance\_6 & -13.55 & -30.05 & -52.35 & --- & -16.30 & -13.85 & -15.25 & -79.15 & -31.85 & -41.35 \\
 & instance\_7 & -15.80 & -18.20 & -15.35 & --- & -128.20 & -15.45 & -14.70 & -44.75 & -38.50 & -46.25 \\
 & instance\_8 & -49.30 & -71.60 & -12.35 & --- & -62.05 & -16.45 & -15.60 & -38.60 & -16.70 & -14.70 \\
 & instance\_9 & -14.80 & -15.15 & -36.90 & --- & -49.60 & -13.35 & -13.20 & -165.45 & -103.50 & -14.80 \\
 & instance\_10 & -36.20 & -24.55 & -56.35 & --- & -136.95 & -20.90 & -13.60 & -14.20 & -16.75 & -14.15 \\
\midrule
Llama & instance\_1 & -138.40 & -37.40 & -109.65 & --- & -56.85 & -16.40 & -13.70 & -13.85 & -15.45 & -19.50 \\
 & instance\_2 & -104.83 & -62.30 & -36.40 & --- & -21.90 & -16.30 & -26.30 & -18.20 & -13.90 & -15.20 \\
 & instance\_3 & -108.60 & -42.45 & -25.40 & --- & -44.20 & -16.75 & -21.80 & -27.65 & -34.25 & -16.20 \\
 & instance\_4 & -129.01 & -192.05 & -22.05 & --- & -26.80 & -20.70 & -23.25 & -15.20 & -24.75 & -89.00 \\
 & instance\_5 & -52.69 & -149.15 & -20.05 & --- & -27.00 & -27.05 & -65.50 & -26.50 & -18.80 & -15.55 \\
 & instance\_6 & -120.97 & -76.20 & -38.75 & --- & -26.30 & -19.85 & -40.05 & -18.85 & -106.70 & -14.50 \\
 & instance\_7 & -125.31 & -15.20 & -33.65 & --- & -32.95 & -29.55 & -17.95 & -19.85 & -12.65 & -14.85 \\
 & instance\_8 & -51.94 & -80.00 & -50.60 & --- & -20.50 & -14.20 & -29.05 & -95.65 & -42.60 & -15.60 \\
 & instance\_9 & -20.41 & -51.10 & -40.45 & --- & -20.50 & -15.30 & -14.90 & -27.25 & -68.35 & -24.60 \\
 & instance\_10 & -73.76 & -119.20 & -30.85 & --- & -54.20 & -15.50 & -73.75 & -15.20 & -32.15 & -14.15 \\
\midrule
Qwen3 & instance\_1 & -24.50 & -28.40 & -18.00 & -14.35 & -14.85 & -16.70 & -212.35 & -36.20 & -17.10 & -16.75 \\
 & instance\_2 & -18.30 & -17.55 & -68.35 & -15.90 & -15.15 & -16.70 & -13.80 & -39.95 & -20.65 & -18.80 \\
 & instance\_3 & -127.70 & -22.60 & -17.50 & -15.80 & -20.40 & -19.00 & -15.00 & -119.55 & -22.60 & -22.90 \\
 & instance\_4 & -25.10 & -21.80 & -15.00 & -15.60 & -19.60 & -32.45 & -14.95 & -46.50 & -15.70 & -18.75 \\
 & instance\_5 & -15.05 & -19.55 & -16.75 & -14.75 & -17.35 & -34.85 & -16.25 & -43.15 & -17.55 & -17.25 \\
 & instance\_6 & -18.10 & -71.65 & -16.35 & -17.35 & -20.65 & -16.05 & -16.60 & -31.10 & -21.60 & -19.75 \\
 & instance\_7 & -52.25 & -22.55 & -14.25 & -22.70 & -16.45 & -21.55 & -15.90 & -45.05 & -24.70 & -20.25 \\
 & instance\_8 & -30.90 & -24.05 & -16.10 & -24.25 & -15.70 & -15.30 & -14.25 & -17.45 & -18.80 & -35.15 \\
 & instance\_9 & -19.90 & -22.20 & -19.80 & -18.70 & -16.45 & -14.10 & -20.75 & -52.10 & -16.15 & -16.70 \\
 & instance\_10 & -14.85 & -19.55 & -14.05 & -14.10 & -13.75 & -15.50 & -15.95 & -47.15 & -19.25 & -20.65 \\
\bottomrule
\end{tabular}
\end{table*}
\paragraph{Summary.} The raw data reveal three key patterns supporting the main text claims: (1) baseline distributions are heavy-tailed with frequent catastrophic failures; (2) FORGE Protocol training substantially compresses this variance and elevates the returns of most instances into the $-10$ to $-50$ range; and (3) Reflexion exhibits higher instability, with several instances regressing to near-baseline performance. These complete data are provided for reproducibility and to enable meta-analyses of representation-specific failure modes.

\section{Per-Run Detailed Results}
\label{app:results}
This section provides session-level summaries for all experiments, aggregating the raw evaluation data from Appendix~\ref{app:raw-scores} into per-session metrics. Table~\ref{tab:full-results-llama}, Table~\ref{tab:full-results-qwen3}, Table~\ref{tab:full-results-gemini} and Table~\ref{tab:full-results-grok} list average return, total token cost, and graduation counts for each training run, enabling direct comparison of resource efficiency across configurations.

\begin{table}[htbp]
\centering
\scriptsize
\caption{Complete Experimental Results: Gemini Flash-Lite}
\label{tab:full-results-gemini}
\begin{tabular}{clllcc}
\toprule
\textbf{\#} & \textbf{Method} & \textbf{Representation} & \textbf{Avg Reward} & \textbf{Tokens} & \textbf{Graduated} \\
\midrule 
1 & FORGE & Examples & -28.73 & 8.3M & 7/10 \\
2 & FORGE & Examples & -18.72 & 8.6M & 9/10 \\
3 & FORGE & Examples & -24.24 & 6.0M & 10/10 \\
4 & FORGE & Examples & -23.18 & 6.7M & 9/10 \\
5 & FORGE & Examples & -36.11 & 9.3M & 8/10 \\
6 & FORGE & Examples & -24.45 & 10.5M & 8/10 \\
7 & FORGE & Examples & -16.01 & 9.4M & 10/10 \\
8 & FORGE & Mixed & -20.04 & 11.4M & 8/10 \\
9 & FORGE & Mixed & -38.67 & 12.6M & 7/10 \\
10 & FORGE & Mixed & -42.06 & 8.3M & 9/10 \\
11 & FORGE & Mixed & -34.37 & 11.4M & 6/10 \\
12 & FORGE & Mixed & -38.46 & 8.4M & 7/10 \\
13 & FORGE & Mixed & -27.43 & 10.1M & 7/10 \\
14 & FORGE & Mixed & -24.36 & 8.8M & 10/10 \\
15 & FORGE & Rules & -19.86 & 4.7M & 10/10 \\
16 & FORGE & Rules & -47.80 & 8.8M & 6/10 \\
17 & FORGE & Rules & -21.76 & 6.2M & 10/10 \\
18 & FORGE & Rules & -16.26 & 8.4M & 7/10 \\
19 & FORGE & Rules & -28.24 & 9.0M & 7/10 \\
20 & FORGE & Rules & -50.20 & 6.2M & 10/10 \\
21 & FORGE & Rules & -29.85 & 7.5M & 8/10 \\
\midrule
22 & Reflexion & Examples & -58.60 & 13.0M & --- \\
23 & Reflexion & Examples & -77.38 & 12.5M & --- \\
24 & Reflexion & Examples & -102.35 & 13.4M & --- \\
25 & Reflexion & Examples & -53.70 & 13.3M & --- \\
26 & Reflexion & Mixed & -63.05 & 16.6M & --- \\
27 & Reflexion & Mixed & -71.89 & 17.8M & --- \\
28 & Reflexion & Mixed & -82.00 & 16.8M & --- \\
29 & Reflexion & Mixed & -110.60 & 16.9M & --- \\
30 & Reflexion & Rules & -34.42 & 11.6M & --- \\
31 & Reflexion & Rules & -48.01 & 11.2M & --- \\
32 & Reflexion & Rules & -96.88 & 11.6M & --- \\
33 & Reflexion & Rules & -66.29 & 12.5M & --- \\
34 & Reflexion & Rules & -84.29 & 11.0M & --- \\
35 & Reflexion & Rules & -77.06 & 11.3M & --- \\
36 & Reflexion & Rules & -31.73 & 10.7M & --- \\
\midrule
37 & Zero-Shot & --- & -189.61 & 0.5M & --- \\
\bottomrule
\end{tabular}
\end{table}

\begin{table}[htbp]
\centering
\scriptsize
\caption{Complete Experimental Results: Grok}
\label{tab:full-results-grok}
\begin{tabular}{clllcc}
\toprule
\textbf{\#} & \textbf{Method} & \textbf{Representation} & \textbf{Avg Reward} & \textbf{Tokens} & \textbf{Graduated} \\
\midrule
1 & FORGE & Examples & -88.39 & 12.2M & 4/10 \\
2 & FORGE & Examples & -22.82 & 4.1M & 10/10 \\
3 & FORGE & Examples & -16.78 & 9.0M & 10/10 \\
4 & FORGE & Mixed & -50.89 & 8.0M & 9/10 \\
5 & FORGE & Mixed & -53.45 & 10.7M & 7/10 \\
6 & FORGE & Mixed & -22.36 & 9.1M & 10/10 \\
7 & FORGE & Rules & -39.37 & 3.7M & 9/10 \\
8 & FORGE & Rules & -30.11 & 4.9M & 8/10 \\
9 & FORGE & Rules & -31.65 & 5.5M & 9/10 \\
\midrule
10 & Reflexion & Examples & -67.09 & 16.3M & --- \\
11 & Reflexion & Examples & -65.44 & 17.8M & --- \\
12 & Reflexion & Examples & -61.76 & 17.8M & --- \\
13 & Reflexion & Mixed & -111.75 & 20.6M & --- \\
14 & Reflexion & Mixed & -118.07 & 19.1M & --- \\
15 & Reflexion & Mixed & -113.24 & 20.1M & --- \\
16 & Reflexion & Rules & -100.22 & 8.6M & --- \\
17 & Reflexion & Rules & -72.60 & 9.4M & --- \\
18 & Reflexion & Rules & -66.95 & 8.4M & --- \\
\midrule
19 & Zero-Shot & --- & -66.11 & 0.5M & --- \\
20 & Zero-Shot & --- & -50.78 & 0.5M & --- \\
\bottomrule
\end{tabular}
\end{table}

\subsection{Learning Dynamics: Per-Stage Checkpoint Progression}
\label{app:stage-progression}

Tables~\ref{tab:stage-checkpoints}--\ref{tab:stage-checkpoints-indiv} report checkpoint returns at each stage (S1-S6) during training, computed by averaging the frozen checkpoint scores across all instances. Under the \textsc{Best} protocol (Table~\ref{tab:stage-checkpoints}), most configurations exhibit progressive improvement or stabilization across stages. For example, Gemini \textsc{Rules} converges from $-93.7$ (S1) to $-27.7$ (S5), while Qwen3 \textsc{Examples} rapidly stabilizes by S2 ($-22.3$). Notable exceptions include Grok \textsc{Rules}, which experiences late-stage regression (S4--S6 $\approx -224$), reflecting rare failure cascades despite earlier success.
\begin{table*}[htbp]
\centering
\caption{Per-Stage Checkpoint Rewards (FORGE, All Models)}
\label{tab:stage-checkpoints}
\scriptsize
\begin{tabular}{llccccccc}
\toprule
\textbf{Model} & \textbf{Representation} & \textbf{S1} & \textbf{S2} & \textbf{S3} & \textbf{S4} & \textbf{S5} & \textbf{S6} & \textbf{Final} \\
\midrule
Gemini & rules & -93.74$\pm$84.8 & -55.16$\pm$63.6 & -43.31$\pm$63.3 & -49.79$\pm$62.8 & -27.73$\pm$40.8 & -43.88$\pm$57.3 & -30.57$\pm$37.0 \\
Gemini & examples & -90.33$\pm$78.9 & -46.01$\pm$51.8 & -28.32$\pm$26.1 & -23.36$\pm$21.6 & -45.20$\pm$65.2 & -27.45$\pm$20.9 & -24.49$\pm$21.1 \\
Gemini & mixed & -97.12$\pm$82.5 & -71.32$\pm$72.2 & -45.83$\pm$53.9 & -62.50$\pm$68.6 & -70.24$\pm$73.4 & -62.10$\pm$73.4 & -32.20$\pm$34.6 \\
\midrule
Grok & rules & -120.10$\pm$79.4 & -42.29$\pm$50.9 & -47.07$\pm$57.6 & -43.39$\pm$61.7 & -48.49$\pm$71.5 & -51.75$\pm$61.1 & -33.71$\pm$32.8 \\
Grok & examples & -88.97$\pm$85.9 & -83.18$\pm$89.6 & -97.10$\pm$92.4 & -57.92$\pm$52.3 & -109.28$\pm$82.4 & -114.13$\pm$60.5 & -42.66$\pm$54.1 \\
Grok & mixed & -139.28$\pm$82.8 & -32.21$\pm$43.8 & -55.71$\pm$72.4 & -47.69$\pm$59.1 & -63.44$\pm$68.3 & -134.65$\pm$59.3 & -42.24$\pm$44.5 \\
\midrule
Llama & rules & -99.43$\pm$83.7 & -86.97$\pm$78.4 & -68.36$\pm$70.2 & -75.38$\pm$69.9 & -78.69$\pm$77.1 & -89.58$\pm$75.6 & -81.34$\pm$76.5 \\
Llama & examples & -73.62$\pm$68.1 & -56.81$\pm$61.6 & -46.77$\pm$41.8 & -44.04$\pm$46.4 & -46.58$\pm$52.2 & -30.71$\pm$26.3 & -28.30$\pm$22.3 \\
Llama & mixed & -67.14$\pm$74.8 & -60.11$\pm$64.6 & -42.92$\pm$32.3 & -27.91$\pm$16.8 & -90.59$\pm$72.0 & -43.40$\pm$68.0 & -29.57$\pm$37.6 \\
\midrule
Qwen3 & rules & -82.60$\pm$86.6 & -70.50$\pm$80.2 & -41.61$\pm$54.6 & -34.70$\pm$46.9 & -37.00$\pm$45.2 & -33.25$\pm$43.2 & -25.16$\pm$30.5 \\
Qwen3 & examples & -49.42$\pm$62.8 & -22.78$\pm$18.5 & -23.43$\pm$27.9 & -35.04$\pm$51.9 & -41.68$\pm$65.2 & -44.86$\pm$79.4 & -24.28$\pm$36.1 \\
Qwen3 & mixed & -74.77$\pm$79.5 & -63.03$\pm$74.9 & -72.81$\pm$78.8 & -45.27$\pm$52.3 & -45.47$\pm$54.5 & -40.29$\pm$37.4 & -29.31$\pm$23.1 \\
\bottomrule
\end{tabular}
\end{table*}

Under the Reflexion baseline (Table~\ref{tab:stage-checkpoints-indiv}), learning trajectories are more erratic. Without champion broadcast, instances often regress between stages (e.g., Gemini Examples degrades from S5 to S6, while Qwen3 Mixed consistently worsens from S2 onward). Comparing the ``Final'' columns across protocols confirms that FORGE achieves systematically better post-session evaluation performance than Reflexion for most configurations, with particularly large gaps for weaker baseline models (Qwen, Llama).

\begin{table*}[htbp]
\centering
\caption{Per-Stage Checkpoint Rewards (Reflexion, All Models)}
\label{tab:stage-checkpoints-indiv}
\scriptsize
\begin{tabular}{llccccccc}
\toprule
\textbf{Model} & \textbf{Representation} & \textbf{S1} & \textbf{S2} & \textbf{S3} & \textbf{S4} & \textbf{S5} & \textbf{S6} & \textbf{Final} \\
\midrule
Gemini & rules & -82.75$\pm$80.3 & -92.69$\pm$84.4 & -74.49$\pm$72.9 & -71.64$\pm$76.2 & -89.30$\pm$136.5 & -67.74$\pm$73.1 & -62.67$\pm$69.2 \\
Gemini & examples & -115.00$\pm$83.4 & -107.55$\pm$77.4 & -85.93$\pm$71.9 & -83.23$\pm$73.5 & -76.66$\pm$74.9 & -89.64$\pm$75.2 & -78.88$\pm$71.6 \\
Gemini & mixed & -116.16$\pm$85.3 & -110.89$\pm$87.1 & -93.21$\pm$83.6 & -106.12$\pm$84.3 & -80.95$\pm$82.1 & -81.47$\pm$81.6 & -81.91$\pm$80.1 \\
\midrule
Grok & rules & -106.42$\pm$89.7 & -93.64$\pm$85.2 & -83.54$\pm$69.1 & -79.27$\pm$72.0 & -91.54$\pm$77.6 & -97.76$\pm$66.5 & -79.92$\pm$73.0 \\
Grok & examples & -85.56$\pm$84.1 & -80.56$\pm$72.7 & -78.12$\pm$69.5 & -67.28$\pm$62.0 & -81.71$\pm$76.1 & -56.60$\pm$59.5 & -64.76$\pm$66.7 \\
Grok & mixed & -107.05$\pm$90.5 & -60.00$\pm$60.9 & -79.84$\pm$69.6 & -120.35$\pm$82.8 & -113.72$\pm$85.4 & -107.11$\pm$79.4 & -114.35$\pm$81.6 \\
\midrule
Llama & rules & -84.84$\pm$78.3 & -100.55$\pm$73.3 & -81.14$\pm$73.9 & -83.23$\pm$73.2 & -95.30$\pm$77.5 & -87.73$\pm$77.6 & -101.42$\pm$77.6 \\
Llama & examples & -80.95$\pm$79.6 & -67.70$\pm$61.1 & -98.51$\pm$78.3 & -63.47$\pm$63.0 & -77.85$\pm$71.9 & -49.66$\pm$62.9 & -53.95$\pm$64.8 \\
Llama & mixed & -60.04$\pm$69.9 & -75.45$\pm$79.5 & -65.17$\pm$68.2 & -55.68$\pm$59.1 & -37.64$\pm$43.1 & -58.96$\pm$61.7 & -44.24$\pm$51.1 \\
\midrule
Qwen3 & rules & -56.77$\pm$65.9 & -68.90$\pm$81.1 & -88.17$\pm$91.4 & -78.50$\pm$90.6 & -84.23$\pm$86.0 & -97.83$\pm$94.5 & -88.38$\pm$86.4 \\
Qwen3 & examples & -76.81$\pm$79.4 & -43.70$\pm$52.0 & -54.92$\pm$72.0 & -52.81$\pm$68.8 & -64.06$\pm$75.0 & -54.48$\pm$73.6 & -57.57$\pm$71.3 \\
Qwen3 & mixed & -72.06$\pm$84.5 & -75.34$\pm$82.5 & -83.01$\pm$90.5 & -90.80$\pm$93.8 & -86.43$\pm$92.6 & -80.61$\pm$89.6 & -80.40$\pm$89.3 \\
\bottomrule
\end{tabular}
\end{table*}

\begin{table*}[htbp]
  \centering
  \scriptsize
  \caption{Raw Evaluation Scores: Gemini Flash-Lite, Reflexion}
  \label{tab:raw-gemini-indiv}
  \begin{tabular}{l ccccccc cccc cccc}
  \toprule
  \textbf{Instance} & \multicolumn{7}{c}{\textbf{Rules}} & \multicolumn{4}{c}{\textbf{Examples}} & \multicolumn{4}{c}{\textbf{Mixed}} \\
  \cmidrule(lr){2-8}\cmidrule(lr){9-12}\cmidrule(lr){13-16}
   & \textbf{S1} & \textbf{S2} & \textbf{S3} & \textbf{S4} & \textbf{S5} & \textbf{S6} & \textbf{S7} & \textbf{S1} & \textbf{S2} & \textbf{S3} & \textbf{S4} & \textbf{S1} & \textbf{S2} & \textbf{S3} & \textbf{S4} \\
  \midrule
  instance\_1 & -13.90 & -17.30 & -210.70 & -210.70 & -181.10 & -47.80 & -13.80 & -25.80 & -28.30 & -88.55 & -94.60 & -15.60 & -54.30 & -33.05 & -92.85 \\
  instance\_2 & -27.35 & -177.90 & -14.85 & -15.75 & -16.80 & -74.45 & -35.15 & -142.50 & -222.55 & -99.50 & -18.35 & -207.50 & -223.35 & -14.40 & -17.80 \\
  instance\_3 & -32.40 & -110.30 & -31.00 & -14.10 & -37.45 & -31.70 & -36.60 & -57.95 & -37.85 & -42.20 & -94.80 & -13.75 & -14.10 & -21.70 & -36.85 \\
  instance\_4 & -15.75 & -13.30 & -211.20 & -29.45 & -45.90 & -15.15 & -16.20 & -65.25 & -15.10 & -161.80 & -33.45 & -177.90 & -106.75 & -183.40 & -176.70 \\
  instance\_5 & -18.90 & -33.85 & -49.80 & -19.70 & -15.20 & -16.45 & -16.35 & -112.75 & -222.70 & -17.40 & -15.95 & -15.80 & -146.55 & -224.20 & -15.35 \\
  instance\_6 & -31.70 & -13.85 & -86.60 & -117.75 & -223.80 & -120.60 & -71.70 & -14.20 & -143.55 & -182.90 & -15.60 & -24.50 & -33.20 & -34.50 & -163.05 \\
  instance\_7 & -15.15 & -29.25 & -128.95 & -118.85 & -14.85 & -80.00 & -16.85 & -36.20 & -34.90 & -118.25 & -24.00 & -13.85 & -38.30 & -30.20 & -215.20 \\
  instance\_8 & -59.45 & -14.25 & -108.50 & -93.00 & -18.60 & -134.90 & -74.80 & -32.25 & -17.55 & -37.25 & -119.30 & -108.50 & -69.95 & -27.75 & -82.10 \\
  instance\_9 & -69.20 & -55.40 & -45.35 & -20.20 & -98.20 & -204.60 & -21.65 & -37.70 & -14.80 & -224.25 & -38.40 & -33.30 & -14.60 & -110.95 & -132.05 \\
  instance\_10 & -60.40 & -14.70 & -81.85 & -23.35 & -191.00 & -45.00 & -14.25 & -61.45 & -36.45 & -117.95 & -82.60 & -19.85 & -17.80 & -100.30 & -174.10 \\
  instance\_11 & --- & --- & --- & --- & --- & --- & --- & --- & --- & -119.00 & --- & --- & --- & -38.70 & --- \\
  instance\_12 & --- & --- & --- & --- & --- & --- & --- & --- & --- & -223.80 & --- & --- & --- & -34.95 & --- \\
  instance\_13 & --- & --- & --- & --- & --- & --- & --- & --- & --- & -59.85 & --- & --- & --- & -30.15 & --- \\
  instance\_14 & --- & --- & --- & --- & --- & --- & --- & --- & --- & -55.25 & --- & --- & --- & -223.80 & --- \\
  instance\_15 & --- & --- & --- & --- & --- & --- & --- & --- & --- & -101.30 & --- & --- & --- & -38.35 & --- \\
  instance\_16 & --- & --- & --- & --- & --- & --- & --- & --- & --- & -82.35 & --- & --- & --- & -223.75 & --- \\
  instance\_17 & --- & --- & --- & --- & --- & --- & --- & --- & --- & -62.35 & --- & --- & --- & -28.80 & --- \\
  instance\_18 & --- & --- & --- & --- & --- & --- & --- & --- & --- & -82.85 & --- & --- & --- & -193.20 & --- \\
  instance\_19 & --- & --- & --- & --- & --- & --- & --- & --- & --- & -74.80 & --- & --- & --- & -14.45 & --- \\
  instance\_20 & --- & --- & --- & --- & --- & --- & --- & --- & --- & -95.40 & --- & --- & --- & -33.40 & --- \\
  \bottomrule
  \end{tabular}
  \end{table*}  

\begin{table*}[htbp]
\centering
\scriptsize
\caption{Raw Evaluation Scores: Reflexion (Grok, Llama, Qwen)}
\label{tab:raw-others-indiv}
\begin{tabular}{ll cccc ccc ccc}
\toprule
\textbf{Model} & \textbf{Instance} & \multicolumn{4}{c}{\textbf{Rules}} & \multicolumn{3}{c}{\textbf{Examples}} & \multicolumn{3}{c}{\textbf{Mixed}} \\
\cmidrule(lr){3-6}\cmidrule(lr){7-9}\cmidrule(lr){10-12}
 & & \textbf{S1} & \textbf{S2} & \textbf{S3} & \textbf{S4} & \textbf{S1} & \textbf{S2} & \textbf{S3} & \textbf{S1} & \textbf{S2} & \textbf{S3} \\
\midrule
Grok & instance\_1 & -15.05 & -20.15 & -13.95 & --- & -19.95 & -97.30 & -206.40 & -53.85 & -196.70 & -91.40 \\
 & instance\_2 & -210.70 & -106.70 & -62.70 & --- & -105.15 & -31.10 & -110.15 & -134.30 & -14.80 & -199.80 \\
 & instance\_3 & -46.50 & -110.70 & -21.55 & --- & -162.45 & -22.70 & -14.85 & -41.90 & -38.50 & -172.30 \\
 & instance\_4 & -98.65 & -15.15 & -58.95 & --- & -76.35 & -14.00 & -86.25 & -26.50 & -139.70 & -174.30 \\
 & instance\_5 & -217.75 & -72.80 & -13.00 & --- & -35.05 & -105.25 & -22.35 & -165.60 & -223.75 & -22.30 \\
 & instance\_6 & -15.40 & -57.65 & -83.60 & --- & -15.60 & -39.40 & -50.75 & -59.70 & -20.95 & -136.35 \\
 & instance\_7 & -28.75 & -45.35 & -72.65 & --- & -24.30 & -39.40 & -61.95 & -41.30 & -202.55 & -29.60 \\
 & instance\_8 & -14.15 & -113.80 & -188.20 & --- & -99.10 & -24.50 & -31.20 & -223.80 & -202.90 & -57.25 \\
 & instance\_9 & -190.45 & -101.80 & -121.40 & --- & -17.75 & -130.25 & -13.05 & -146.75 & -54.20 & -224.30 \\
 & instance\_10 & -164.75 & -81.90 & -33.45 & --- & -115.20 & -150.45 & -20.65 & -223.80 & -86.60 & -24.80 \\
\midrule
Llama & instance\_1 & -16.20 & -63.20 & -113.60 & -27.00 & -14.50 & -15.05 & --- & -13.45 & -26.40 & --- \\
 & instance\_2 & -71.50 & -132.30 & -90.80 & -23.60 & -19.10 & -14.00 & --- & -132.30 & -93.05 & --- \\
 & instance\_3 & -29.40 & -166.55 & -196.30 & -114.10 & -17.25 & -37.35 & --- & -16.55 & -103.70 & --- \\
 & instance\_4 & -43.25 & -22.55 & -159.45 & -24.80 & -51.95 & -16.05 & --- & -17.15 & -16.35 & --- \\
 & instance\_5 & -108.60 & -56.20 & -218.70 & -143.60 & -14.15 & -122.35 & --- & -69.95 & -12.80 & --- \\
 & instance\_6 & -128.15 & -149.80 & -125.20 & -168.60 & -36.05 & -18.95 & --- & -18.05 & -36.20 & --- \\
 & instance\_7 & -29.10 & -174.05 & -121.75 & -225.65 & -53.55 & -89.30 & --- & -135.40 & -42.15 & --- \\
 & instance\_8 & -173.35 & -68.20 & -34.45 & -216.35 & -48.30 & -224.30 & --- & -38.35 & -14.55 & --- \\
 & instance\_9 & -18.85 & -37.15 & -112.05 & -127.50 & -31.00 & -106.20 & --- & -27.95 & -22.55 & --- \\
 & instance\_10 & -14.80 & -199.35 & -36.95 & -125.75 & -24.90 & -13.35 & --- & -13.95 & -23.30 & --- \\
 & instance\_11 & --- & -138.05 & -28.35 & -96.45 & -20.10 & --- & --- & -20.30 & --- & --- \\
 & instance\_12 & --- & -122.20 & -80.85 & -157.10 & -31.05 & --- & --- & -119.75 & --- & --- \\
 & instance\_13 & --- & -224.15 & -40.25 & -82.75 & -179.70 & --- & --- & -15.55 & --- & --- \\
 & instance\_14 & --- & -129.80 & -66.55 & -148.95 & -16.75 & --- & --- & -22.65 & --- & --- \\
 & instance\_15 & --- & -28.10 & -49.45 & -54.85 & -17.00 & --- & --- & -29.70 & --- & --- \\
 & instance\_16 & --- & -106.70 & -35.60 & -136.75 & -30.35 & --- & --- & -19.70 & --- & --- \\
 & instance\_17 & --- & -75.10 & -83.90 & -210.40 & -223.80 & --- & --- & -13.90 & --- & --- \\
 & instance\_18 & --- & -89.50 & -43.25 & -32.15 & -34.00 & --- & --- & -122.25 & --- & --- \\
 & instance\_19 & --- & -125.00 & -33.80 & -137.95 & -62.70 & --- & --- & -72.25 & --- & --- \\
 & instance\_20 & --- & -165.75 & -92.50 & -174.10 & -35.25 & --- & --- & -17.00 & --- & --- \\
\midrule
Qwen3 & instance\_1 & -18.90 & -20.30 & --- & --- & -22.65 & -17.60 & --- & -30.90 & -212.20 & --- \\
 & instance\_2 & -37.95 & -26.70 & --- & --- & -20.60 & -14.15 & --- & -223.70 & -223.75 & --- \\
 & instance\_3 & -95.25 & -132.00 & --- & --- & -16.05 & -16.00 & --- & -15.10 & -72.25 & --- \\
 & instance\_4 & -23.55 & -21.30 & --- & --- & -184.95 & -14.00 & --- & -224.35 & -19.35 & --- \\
 & instance\_5 & -16.95 & -85.15 & --- & --- & -21.65 & -16.35 & --- & -15.65 & -18.00 & --- \\
 & instance\_6 & -16.85 & -16.35 & --- & --- & -115.85 & -192.70 & --- & -47.90 & -20.65 & --- \\
 & instance\_7 & -182.75 & -223.75 & --- & --- & -23.70 & -21.30 & --- & -211.70 & -29.65 & --- \\
 & instance\_8 & -223.80 & -186.35 & --- & --- & -17.20 & -15.75 & --- & -58.35 & -210.70 & --- \\
 & instance\_9 & -223.75 & -15.55 & --- & --- & -176.30 & -224.30 & --- & -24.50 & -223.80 & --- \\
 & instance\_10 & -22.20 & -223.80 & --- & --- & -41.35 & -72.00 & --- & -40.85 & -223.70 & --- \\
 & instance\_11 & --- & -20.05 & --- & --- & -148.60 & --- & --- & -17.75 & --- & --- \\
 & instance\_12 & --- & -21.20 & --- & --- & -15.40 & --- & --- & -223.80 & --- & --- \\
 & instance\_13 & --- & -14.55 & --- & --- & -16.65 & --- & --- & -25.50 & --- & --- \\
 & instance\_14 & --- & -204.70 & --- & --- & -18.10 & --- & --- & -19.70 & --- & --- \\
 & instance\_15 & --- & -68.70 & --- & --- & -16.35 & --- & --- & -24.50 & --- & --- \\
 & instance\_16 & --- & -23.15 & --- & --- & -19.75 & --- & --- & -16.55 & --- & --- \\
 & instance\_17 & --- & -95.40 & --- & --- & -17.05 & --- & --- & -224.30 & --- & --- \\
 & instance\_18 & --- & -223.75 & --- & --- & -21.35 & --- & --- & -223.75 & --- & --- \\
 & instance\_19 & --- & -22.65 & --- & --- & -193.65 & --- & --- & -58.25 & --- & --- \\
 & instance\_20 & --- & -144.15 & --- & --- & -15.75 & --- & --- & -18.30 & --- & --- \\
 & instance\_21 & --- & --- & --- & --- & --- & --- & --- & -19.20 & --- & --- \\
 & instance\_22 & --- & --- & --- & --- & --- & --- & --- & -14.35 & --- & --- \\
 & instance\_23 & --- & --- & --- & --- & --- & --- & --- & -18.90 & --- & --- \\
 & instance\_24 & --- & --- & --- & --- & --- & --- & --- & -18.95 & --- & --- \\
 & instance\_25 & --- & --- & --- & --- & --- & --- & --- & -25.45 & --- & --- \\
 & instance\_26 & --- & --- & --- & --- & --- & --- & --- & -22.60 & --- & --- \\
 & instance\_27 & --- & --- & --- & --- & --- & --- & --- & -20.75 & --- & --- \\
 & instance\_28 & --- & --- & --- & --- & --- & --- & --- & -24.05 & --- & --- \\
 & instance\_29 & --- & --- & --- & --- & --- & --- & --- & -21.35 & --- & --- \\
 & instance\_30 & --- & --- & --- & --- & --- & --- & --- & -41.40 & --- & --- \\
 & instance\_31 & --- & --- & --- & --- & --- & --- & --- & -223.60 & --- & --- \\
 & instance\_32 & --- & --- & --- & --- & --- & --- & --- & -15.25 & --- & --- \\
 & instance\_33 & --- & --- & --- & --- & --- & --- & --- & -223.80 & --- & --- \\
 & instance\_34 & --- & --- & --- & --- & --- & --- & --- & -14.20 & --- & --- \\
 & instance\_35 & --- & --- & --- & --- & --- & --- & --- & -21.20 & --- & --- \\
 & instance\_36 & --- & --- & --- & --- & --- & --- & --- & -222.15 & --- & --- \\
 & instance\_37 & --- & --- & --- & --- & --- & --- & --- & -19.65 & --- & --- \\
 & instance\_38 & --- & --- & --- & --- & --- & --- & --- & -19.05 & --- & --- \\
 & instance\_39 & --- & --- & --- & --- & --- & --- & --- & -16.30 & --- & --- \\
 & instance\_40 & --- & --- & --- & --- & --- & --- & --- & -18.45 & --- & --- \\
\bottomrule
\end{tabular}
\end{table*}
  
\section{Reflexion Baseline Raw Scores}
\label{app:individual-scores}
Tables~\ref{tab:raw-gemini-indiv}--\ref{tab:raw-others-indiv} present evaluation scores under the Reflexion baseline, where each instance evolves memory from its own trajectories without cross-instance propagation. 
\paragraph{Gemini Reflexion.} Table~\ref{tab:raw-gemini-indiv} reports Gemini scores under the Reflexion baseline where cross-instance transfer is disabled. Each of the three memory representations (Rules/Examples/Mixed) received 2 evaluation runs.% ---- end appendix/appendix ----

\end{document}